\documentclass{article}

\usepackage[preprint]{neurips_2026}
\usepackage[utf8]{inputenc}
\usepackage[T1]{fontenc}
\usepackage{hyperref}
\usepackage{url}
\usepackage{booktabs}
\usepackage{multirow}
\usepackage{amsfonts}
\usepackage{amsmath}
\usepackage{amssymb}
\usepackage{nicefrac}
\usepackage{microtype}
\usepackage{xcolor}
\usepackage{graphicx}
\usepackage{pdfpages}
\usepackage{subcaption}
\usepackage{algorithm}
\usepackage{algpseudocode}
\usepackage{listings}
\usepackage{enumitem}
\usepackage{array}
\usepackage{tabularx}
\usepackage{placeins}
\usepackage{float}
\usepackage{tikz}
\usetikzlibrary{arrows.meta, positioning, shapes.geometric, fit, backgrounds, calc}

\title{Workspace Optimization:\\ How to Train Your Agent}

\author{%
  Elad Sarafian\thanks{Equal contribution.} \\
  NVIDIA \\
  \texttt{esarafian@nvidia.com} \\
  \And
  Gal Kaplun\footnotemark[1] \\
  NVIDIA \\
  \texttt{gkaplun@nvidia.com} \\
  \And
  Ron Banner \\
  NVIDIA \\
  \texttt{rbanner@nvidia.com} \\
  \And
  Daniel Soudry \\
  NVIDIA \& Technion \\
  \texttt{daniel.soudry@gmail.com} \\
  \And
  Boris Ginsburg \\
  NVIDIA \\
  \texttt{bginsburg@nvidia.com} \\
}

\begin{document}

\maketitle

% ── Main paper (9 pages) ─────────────────────────────────────────────
\begin{abstract}
Modern agents built on frontier language models often cannot adapt their weights. What, then, remains trainable? We argue it is the agent's \emph{workspace}, the structured external substrate it reads, writes, and tests; we call its evolution \emph{workspace optimization}. Workspace optimization targets hard multi-turn environments where a frontier model has strong priors but cannot solve the task in a single shot, so the agent must learn through interaction. We propose a principled way to evolve the workspace, mirroring the structure of weight-space training: artifacts in place of parameters, evidence in place of data, counterexamples in place of losses, and textual feedback in place of gradients. We instantiate the idea in \textsc{DreamTeam}, a multi-agent harness for ARC-AGI-3 whose roles build an executable world model, plan, hypothesize, probe, strategize, and route failures. On the current 25-game ARC-AGI-3 public set under the official scoring protocol and averaged over two independent runs, \textsc{DreamTeam} improves the SOTA protocol-matched agent's score from 36\% to 38.4\%, while using 31\% fewer environment actions per game. See ARC Prize scorecard record (single run): \url{https://arcprize.org/scorecards/831c83cf-b969-45fc-a6ce-27f9b3c4105c}.
\end{abstract}

\section{Introduction}
\label{sec:intro}

Machine learning usually explains adaptation as movement in weight space. Data produces a loss, the loss produces an update, and enough updates change the model. That recipe shapes the default story for agentic learning: when the system fails, collect experience and train the model.

Many deployed agents~\citep{anthropic_claude_code,openai_codex} cannot follow that story. Fine-tuning is too slow for an online loop in which the agent must adapt within a single episode, and in many settings weight access is not on the table to begin with: the model sits behind an API, or the traces a deployed agent generates contain content that should not be baked into a checkpoint. The practical situation is that the model is fixed and the agent still has to learn. What, then, is the trainable state?

Imagine dropping a person into an unfamiliar video game with no instructions. The buttons do something, the screen changes in response, and some of the objects on screen turn out to matter while others are decoration. The goal is not stated (this is the setting of the ARC-AGI-3 benchmark~\citep{arcagi3}). Most people figure things out anyway by reasoning from prior similar experiences: they press buttons, watch what changes, form a rough theory, and start playing on purpose. The task, the environment, and a workable plan all come out of the same interaction.

In standard formulations, the input specifies the problem. In such an unfamiliar setting, the specification is hidden inside the environment instead. The question is not only \emph{what action should the agent take?} It is \emph{what should the agent learn next, where should it store that learning, and how should it change later behavior?} For example, two agents reading the same history can still end up in different places. Consider a door that opens when the player steps on a blue tile. One agent treats the surprise as a counterexample, attributes the failure to its transition rule, and patches the rule. The other appends a note for later retrieval. Both share the same frozen weights; they differ in the state they keep around the model.

We call this state the \emph{workspace}, after the global-workspace framing in cognitive science~\citep{baars2005}. It is the structured material the agent reads, writes, and tests around its calls to the model: some pieces are durable across the episode, some pieces are local to a single step (the parse the agent just did, the probe it is considering). All of it is editable. A workspace is not merely memory; it is the machinery that decides which observations turn into state, which predictions get logged as commitments, and which surprises change future behavior.

This paper studies \emph{workspace optimization}: adapting a frozen-model agent by editing the workspace instead of the weights. The agent acts through the workspace, compares what it expected with~what happened, and changes the state that later calls will read. The interesting question is credit assignment: when something goes wrong, what gets fixed? Different failures point at different parts of the workspace: misreading what is happening and mispredicting what will happen call for different fixes. %Each surprise lands at one of those addresses as a counterexample, rather than disappear into the log.

We instantiate this idea in \textsc{DreamTeam}, a multi-agent orchestrator for unfamiliar interactive games. Its workspace contains executable artifacts for observation, dynamics, and strategy, which the agents read, edit, and test around each action step. Each action is treated as a test of the workspace's predictions: a confirmed prediction is supporting evidence for the responsible artifacts, and a contradicted one is a counterexample routed to the artifact most responsible for the failure. The proposed repair is then replayed against earlier evidence to catch regressions before the next step. Early in a game, the loop favors probes that reduce uncertainty; as World-Model (WM) checks improve, control shifts toward model-based planning and efficient execution against the WM.

\begin{figure}[t]
  \centering
  \begin{subfigure}[b]{0.45\linewidth}
    \centering
    \includegraphics[width=\linewidth]{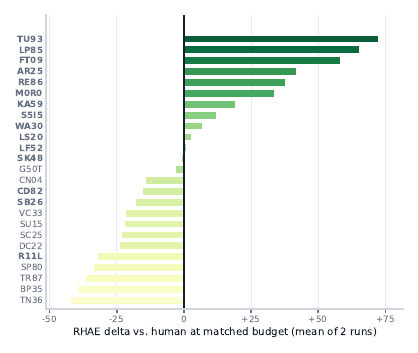}
    \caption{}
    \label{fig:teaser-results}
  \end{subfigure}\hfill
  \begin{subfigure}[b]{0.45\linewidth}
    \centering
    \includegraphics[width=\linewidth]{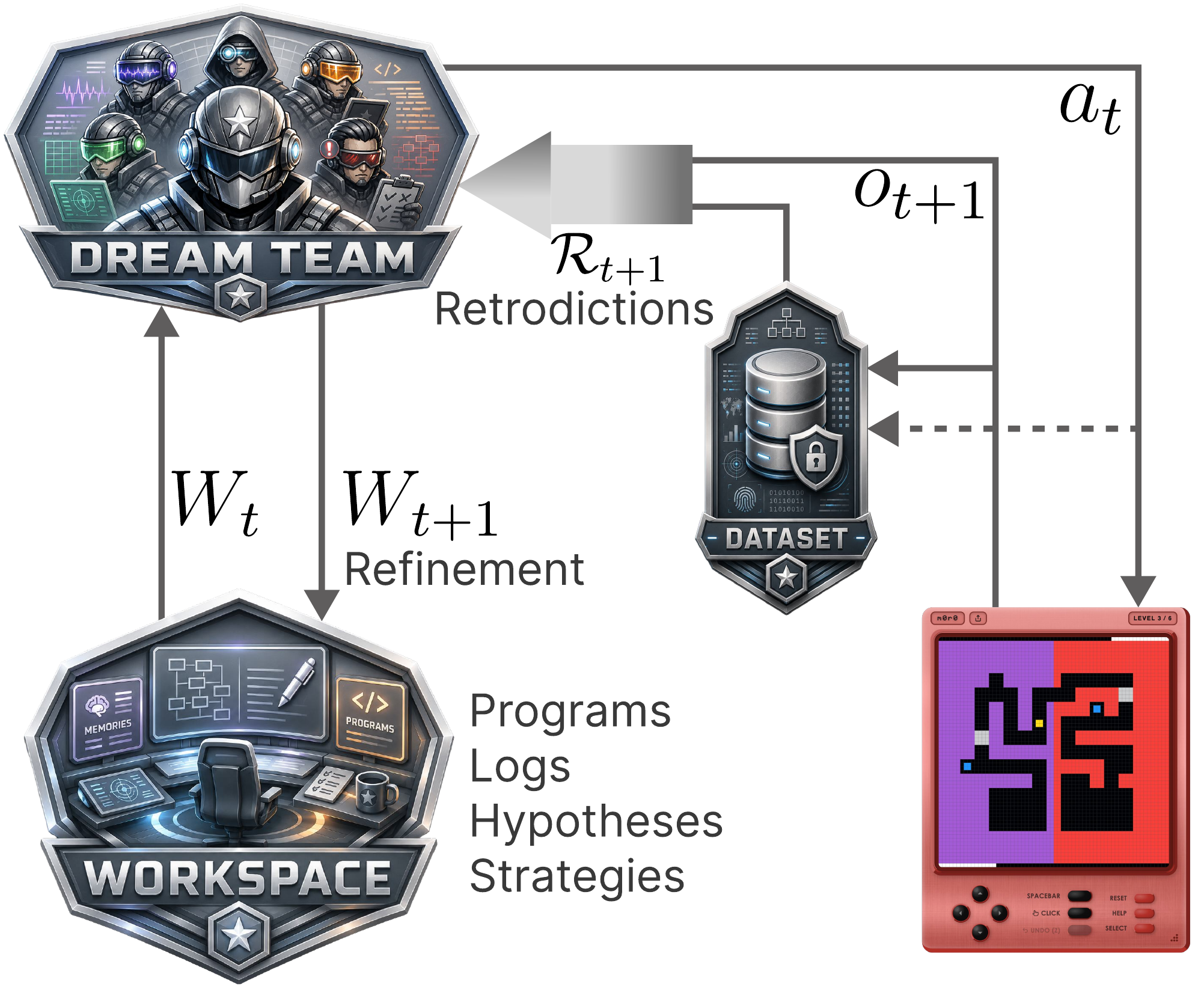}
    \caption{}
    \label{fig:teaser-arch}
  \end{subfigure}
  \caption{%
    \textbf{(a)} Per-game RHAE gap
    $\Delta_g = \mathrm{RHAE}_{\textsc{DreamTeam}}(g) - \mathrm{RHAE}_{\mathrm{human}}(g)$,
    averaged over our two independent runs of \textsc{DreamTeam}, with
    the human score sampled at \textsc{DreamTeam}'s action budget;
    bold game labels mark equal-level finishers.
    \textbf{(b)} The \textsc{DreamTeam} workspace-optimization loop:
    retrodiction over historical evidence yields an updated workspace
    (hypotheses, strategies, logs) that roles read on the next step
    to refine their decisions.
  }
  \label{fig:dreamteam-teaser}
  \vspace{-0.4cm}
\end{figure}

\noindent\textbf{Contributions.} This paper makes three contributions.
\begin{itemize}[leftmargin=*, nosep]
    \item \textbf{Workspace optimization.} We formulate workspace
    optimization, a framework in which frozen-model agents adapt by
    editing a structured workspace of typed artifacts. Each surprise
    is routed to the artifact responsible, recasting credit assignment
    as which artifact to edit rather than how much to update a weight
    (Section~\ref{sec:workspace-optimization}).
    \item \textbf{\textsc{DreamTeam}.} We instantiate workspace
    optimization as \textsc{DreamTeam}, a DreamerV3-inspired~\citep{dreamerv3}
    multi-agent harness whose roles own observation, dynamics,
    strategy, probing, critique, and arbitration, each acting through
    and updating its part of the workspace from contradicted
    predictions (Section~\ref{sec:dreamteam}).
    \item \textbf{Efficient new state of the art.} On the 25-game
    ARC-AGI-3 public set and averaged over two independent runs,
    \textsc{DreamTeam} reaches 38.4\% RHAE (Relative Human Action
    Efficiency, the official ARC-AGI-3 score; full definition in
    Section~\ref{sec:experiments}), surpassing the prior
    protocol-matched single-run baseline of 36\%, while using
    31\% fewer environment actions per game (mean of 444 vs the
    baseline's 643)
    (Figure~\ref{fig:dreamteam-teaser}, Section~\ref{sec:experiments}).
\end{itemize}
\FloatBarrier

\noindent\textbf{Related work.}
Model-based RL and neural World-Models learn predictive state and dynamics in parameter space and plan with imagined rollouts~\citep{worldmodels2018,dreamerv2,dreamerv3,muzero}. Program synthesis and executable-world-model work replaces the latent state with code that humans can read and tests can falsify~\citep{dreamcoder,grand2024lilo,romera2024funsearch,guan2023llmp,hao2023rap,wong2023wordtoworld,webdreamer,worldllm,wang2024hypothesis}. Workspace optimization keeps the computational shape of model-based RL but trains a structured workspace of executable artifacts in place of a parameter vector, with feedback delivered as a counterexample owned by a named artifact rather than a gradient.

A recent line of work adapts at inference time by editing structured external state under frozen weights, with case banks~\citep{zhou2025memento}, distilled strategy libraries~\citep{ouyang2025reasoningbank}, persistent cheatsheets~\citep{suzgun2025cheatsheet}, and self-rewriting agent code~\citep{zhang2025dgm} extending an older text-memory and skill-library line~\citep{react,shinn2023reflexion,packer2023memgpt,voyager,park2023generative} that wrote into less-structured stores at trajectory or session granularity. \textsc{DreamTeam} differs on two structural axes. Failures are routed to one of a small set of typed slots whose owner is named at write time, instead of being retrieved by similarity, which turns credit assignment into an addressing decision and sidesteps the open failure-attribution problem~\citep{zhang2025attribution}. Each accepted patch is replayed against earlier transitions retained as a regression set, so a revision must remain consistent with prior evidence as well as the current outcome. We give detailed related work in Appendix~\ref{app:extended-related-work}.\vspace{-0.05cm}

\section{Workspace Optimization}
\label{sec:workspace-optimization}
\label{sec:workspace}
This section introduces workspace optimization, in which a frozen model accumulates task-specific competence by editing a mutable workspace of programs, traces, and logs while a harness replays earlier transitions to evaluate each edit.\vspace{-0.1cm}

\begin{table}[h]
\centering
\caption{Persistent workspace components shaped by experience and read in future rounds.}
\label{tab:workspace-lifecycle-v6}
\small
\renewcommand{\arraystretch}{1.15}
\begin{tabularx}{\linewidth}{@{}>{\centering\arraybackslash}p{0.06\linewidth}>{\raggedright\arraybackslash}p{0.12\linewidth}>{\raggedright\arraybackslash}X>{\raggedright\arraybackslash}X@{}}
\toprule
\textbf{Symbol} & \textbf{Name} & \textbf{What it is} & \textbf{Update effect} \\
\midrule
\(\mathcal{P}_t\) & Programs & Reusable procedures, policies, renderers, or other callable artifacts. & Edits the computations for future calls. \\
\addlinespace[0.2em]
\(\mathcal{L}_t\) & Logs & Summaries, hypotheses, abandoned candidates, and unresolved questions. & Updates the agent's interpretive context. \\
\(\mathcal{D}_t\) & Trace data & Observations, actions, rewards, and counterexamples. & Appends new evidence from interaction. \\
% \addlinespace[0.1em]
\bottomrule
\end{tabularx}
\end{table}

\vspace{-0.08cm}
\subsection{The Trainable State}

  % We model an agent as a tuple \((M, H)\) of a  language model
  % \(M\) and a harness \(H\) providing tools, a sandbox, and a context
  % manager. In an interactive, multi-turn environment, the harness
  % exposes a structured external state that the agent reads, writes, and
  % tests around each call to \(M\). Formally, for step $t$ we associate
  % with the agent a \textbf{mutable workspace}

 We model an agent as a tuple \((M, H)\) of a language model \(M\) and a harness \(H\) providing tools, a sandbox, and a context manager. In an interactive, multi-turn environment, the
  harness exposes an observation dataset \(\mathcal{D}_t\) and a structured \textbf{mutable workspace} \(W_t\) that the agent reads,
  writes, and tests around each call to \(M\) and each environment interaction. Formally, for step \(t\):
\[
  W_t = (\mathcal{P}_t, \mathcal{L}_t),
  \qquad
  \mathcal{D}_t = (o_0, a_0, r_0, \ldots, o_{t-1}, a_{t-1}, r_{t-1}, o_t).
\]

% {
% \setlength{\abovecaptionskip}{0.1pt}
% \setlength{\belowcaptionskip}{0.1pt}

% }

Standard agent memory is append-only: it records \emph{what has happened} but holds no theory of what comes next, and no rules or code committed for reuse. The workspace and the observation dataset together expose three slots that the agent reads (Table~\ref{tab:workspace-lifecycle-v6}): the append-only trace in \(\mathcal{D}_t\), working hypotheses and plans in \(\mathcal{L}_t\), and programs in \(\mathcal{P}_t\). These components evolve across a run. At step \(t\), the agent reads \(W_t\) to choose an action, evaluates the outcome once the next observation
  \(o_{t+1}\) arrives, and writes the result back as a workspace edit:
  \[
  a_t = \operatorname{Act}(W_t, \mathcal{D}_t),
  \qquad
  e_t = \operatorname{Evaluate}(W_t, \mathcal{D}_{t+1}),
  \qquad
  W_{t+1} = \operatorname{Update}(W_t, e_t).
  \]

% We describe \(\operatorname{Evaluate}\) below and defer \(\operatorname{Update}\) to Section~\ref{sec:comp-graph}.

\noindent\textbf{Evaluation.}
\label{sec:repair-objective}
The signal \(e_t\) comes from three sources.
First, the \emph{external reward} \(r_t\), which is independent of
\(W_t\).
Second, \emph{prediction failure}: a hypothesis in
\(\mathcal{L}_t\) or a rule in \(\mathcal{P}_t\) predicted \(o_{t+1}\), and
the next observation contradicts it. Third, \emph{regression breakage}: the
harness maintains a \emph{regression set} \(\mathcal{R}_t\), a window of
past transitions drawn from \(\mathcal{D}_t\), and replays the just-edited
programs against every entry. A transition the previous version handled
but the patched version no longer handles becomes a new counterexample
for the responsible role on the next step.

\subsection{The Workspace as a Computational Graph}
\label{sec:comp-graph}

\begin{figure}[h]
\centering
\includegraphics[width=\linewidth]{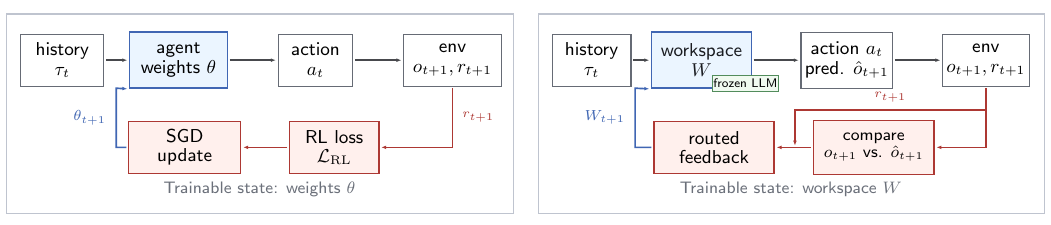}
\vspace{-0.35em}
\begin{minipage}[t]{0.48\linewidth}
\centering\footnotesize\textbf{(a) Weight-space training}
\end{minipage}\hfill
\begin{minipage}[t]{0.48\linewidth}
\centering\footnotesize\textbf{(b) Workspace optimization}
\end{minipage}
\vspace{0.25em}
\caption{Weight-space RL updates agent weights from interaction losses over histories, actions, environment outcomes, and rewards. Workspace optimization keeps the model frozen and updates the external state that future inference reads: artifacts, evidence, summaries, and policies. After each edit, the harness replays earlier transitions through the updated artifacts and reports the deltas as feedback.}
\label{fig:weight-vs-workspace}
\end{figure}

So far we have the nodes of the workspace graph, the components of \(W_t\), and the signal \(e_t\) that scores them, the analogues of weights and loss in standard training. What remains are the edges: how  feedback at one node becomes an edit at another.

In weight-space training (Figure~\ref{fig:weight-vs-workspace}A), the chain rule distributes a scalar loss among the parameters that produced the prediction.
Workspace optimization (Figure~\ref{fig:weight-vs-workspace}) does not need this machinery to name the interface that emitted a failed claim. Each prediction is emitted by one specific interface, so the prediction-failure component of \(e_t\) carries that interface's name. Interface-level attribution is therefore direct; the broader credit-assignment problem is the cascade of repairs that follows once a patch is applied.
The cost of this localization, paid in advance, is committing to interfaces precise enough to be falsified.

Local attribution does not imply local repair: a patch to one interface
can break earlier transitions that depend on it. Suppose a movement rule
predicts that LEFT moves the avatar one cell, a wall blocks the move at
step $t$, and the Simulator patches the rule; on replay against
\(\mathcal{R}_t\), the patch fails an earlier transition where LEFT did
move the avatar, implicating the wall-collision rule that distinguishes
the two cases. Repairs sequence themselves through replay failures,
with no backward operator constructing the path in advance.

Two mechanisms substitute for differentiability. The regression set
\(\mathcal{R}_t\) keeps a single repair from settling on the most recent
transition at the expense of older ones. The workspace's dependency
structure bounds where a repair can land: each edit propagates only
through calls of the edited interface, and inputs the agent cannot
rewrite (the model's weights, the environment, and historical
observations) terminate the chain where stop-gradient would terminate a
backward pass.

\textbf{Remark 1 (Initialization).} Workspace optimization begins from a seed workspace $W_0$, analogous to weight initialization in a neural network. Appendix~\ref{app:artifact-schemas} describes the seed artifacts used in \textsc{DreamTeam}.
The seed sets an inductive bias for how the workspace can evolve. An over-specified seed can constrain exploration and make early repairs overfit its initial ontology. An under-specified seed can slow bootstrapping and leave early behavior imprecise. A good seed gives the harness enough structure to make evidence usable while leaving room for the workspace to change.

\textbf{Remark 2 (Inductive and transductive outputs).} Two agents can run the same model and still differ in what they return: a value, or a callable artifact. We call the first \emph{transductive} and the second \emph{inductive}~\citep{vapnik1998statistical,chapelle1999transductive}. A transductive output (a prediction, action proposal, explanation, or plan) enters \(\mathcal{L}_t\) as context. An inductive output (a rule, renderer, or policy) enters \(\mathcal{P}_t\) and can be invoked on inputs the agent has not seen.\footnote{\textbf{Hybrid agents.} Some agents are both: an observer encodes ``transductively'' but also returns a callable renderer. (Table~\ref{tab:roles})} The inductive choice is preferable when the mapping can be written as code: the function is cheap and falsifiable, at the cost of committing to a fixed interface on potentially scant evidence. The transductive choice is preferable when the input is unstructured or the answer requires the model's judgment, at the cost of less reliability and a tendency toward self-agreement.
        % Workspace Optimization
\section{ARC-AGI-3 Solver: Deriving the Architecture from First Principles}
\label{sec:unknown-game-principles}

ARC-AGI-3 places a frozen language model in front of a game whose rules, mechanics, and goal are unknown. In each step $t$, the agent receives only a transient observation $o_t$ and must select the next action $a_t$. Two features distinguish this setting from standard reinforcement-learning benchmarks. There is no offline interaction data and there is no way to rewind a step or fan out parallel rollouts to fit a policy from. They apply within a tight action-efficiency constraint: human players solve these games in a few hundred moves. Within that budget, every action is an exploration-exploitation trade-off between probing and progressing.

The frozen model brings strong arcade-and-video-game priors about what objects mean, what counts as progress, and what the goal is likely to be. When these priors align with the game's mechanics and goal they make few-hundred-action competence achievable; when they mislead, every subsequent action reinforces the wrong reading in a positive-feedback loop.

\subsection{The Limits of Standard Architectures}

A model-free reinforcement-learning approach has no traction here. The critic starts uninitialized with no signal to guide the policy, so progress depends on exploration alone. Even once exploration clears a level, two problems prevent the standard policy-gradient loop from closing: the trajectory is strictly online and visits each state at most once, so each rewarded transition is a single noisy example, far too weak to fit a critic; and the agent cannot return there to re-optimize, so any improvement has to transfer to states in later levels, a signal that is itself weak.

A pure model-based approach is closer to what the setting demands, since a WM can in principle be queried before each action and updated from each transition. But conventional latent-dynamics models presuppose either offline data or a long online training phase. Even state-of-the-art sample-efficient agents consume thousands of environment steps before their WMs become trustworthy, while ARC-AGI-3 expects competence after dozens. Whatever WM the agent uses must therefore be representable in a form the agent can write down, inspect, and revise.

Both routes leave a deeper problem unaddressed: without a goal hypothesis, exploration has no direction, and within a few hundred actions undirected exploration neither stumbles onto a reward signal nor covers the state space well enough for dynamics learning to converge. How can the agent acquire a goal and explore efficiently inside this budget?

\subsection{The Architecture That Follows}

These observations point in the same direction. The agent needs an explicit, inspectable theory of the game, constructed from interaction and tested by interaction. The theory must be detailed enough to make predictions, so that failures are diagnostic, and structured enough to be repaired locally, so that a single contradicted prediction does not require rebuilding everything.

We call this object the executable game WM and require three properties: a committed prediction before each action, so that retrodicting against the next observation either confirms or contradicts it; identifiability of the offending interface (observation parsing, hidden-state tracking, or action effect) when a prediction fails, so repairs are local; and usability inside the agent for planning once predictions are reliable, so the agent reasons in imagined rollouts as well as real actions.

This is recognizably the Dreamer template from model-based reinforcement learning \citep{dreamerv3}, with one substantive change: the WM is code rather than parameters. We call each piece of the workspace a \emph{surface}: a typed slot owned by one role, holding the artifacts that role reads and edits (six in total, listed in Table~\ref{tab:roles}). The training loop keeps the gradient template, with per-step error signals routed back to whichever surface emitted the failed claim. Edits patch the program in place, and the regression set $\mathcal{R}_t$ replays earlier transitions to evaluate each edit.

\begin{figure}[h!]
  \centering
  \includegraphics[width=\linewidth]{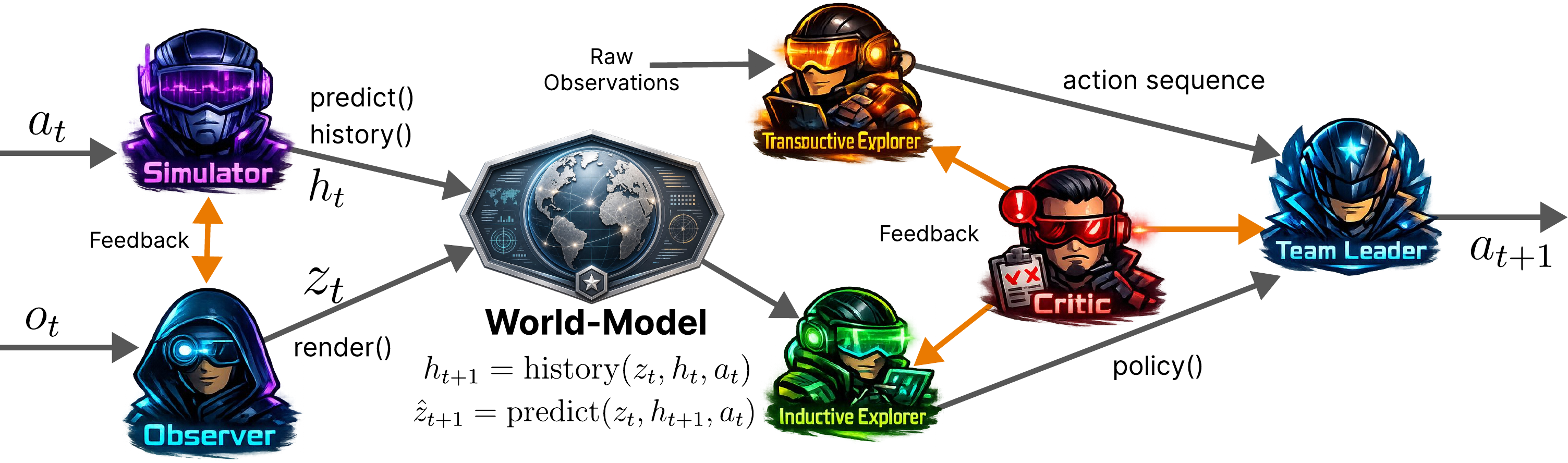}
  \caption{The \textsc{DreamTeam} role graph and executable
  WM interface. The Observer encodes $o_t$ into $z_t$ and grounds
  it via $\operatorname{render}$. The Simulator carries history
  $h_{t+1} = \operatorname{history}(z_t, h_t, a_t)$ and predicts one
  step ahead $\hat z_{t+1} = \operatorname{predict}(z_t, h_{t+1}, a_t)$;
  Observer and Simulator exchange feedback when retrodiction
  disagrees with the next encoding ($\hat z_{t+1} \not\approx z_{t+1}$).
  The Inductive Explorer (IE) proposes a sub-goal set $\mathcal{G}_t$ and policies $\Pi_t$;
  the Transductive Explorer (TE) proposes action-sequence probes. The Critic emits feedback
  (orange edges) to the proposers and to the Team Leader (TL), who commits the
  next action $a_t$.}
  \label{fig:dream-team-wm}

  \vspace{-0.4cm}
\end{figure} 

The remaining question, sketched in Figure~\ref{fig:dream-team-wm}, is who edits the WM and who uses it, and at what granularity. Too coarse: a single role carries observation, dynamics, planning, and arbitration at once, raising per-call cognitive load on every surface. Too fine: inter-role hand-offs and audits eat the per-step wall-clock budget that should pay for env actions. We settle on six roles split along two axes, maintaining the WM versus acting through it. The Observer parses observations into structured state and the Simulator tracks hidden state and patches the dynamics rule when a prediction fails; the Inductive Explorer commits reusable strategies, the Transductive Explorer proposes information-seeking probes, the Critic routes failures to their owners, and the Team Leader arbitrates which proposal is committed. The Transductive Explorer is added beyond the Dreamer template: when the WM is still too immature for rollouts to be informative, a probe-proposing role keeps early exploration directed. %Section~\ref{sec:dreamteam} specifies the surfaces, artifacts, and interfaces.

\section{\textsc{DreamTeam}: Executable Workspace Optimization}
\label{sec:dreamteam}

% \textsc{DreamTeam} instantiates the architecture of Section~\ref{sec:unknown-game-principles} as a working multi-agent orchestrator whose editable workspace combines executable code with persistent role context. 

\subsection{The Agentic Workspace}
\label{sec:agentic-workspace}

\textsc{DreamTeam} materializes the six-role decomposition of Section~\ref{sec:unknown-game-principles} as the team's editable workspace. Each role owns exactly one \emph{surface}: a typed slot holding the artifacts the role reads, edits, and is held responsible for. At the center are the two WM surfaces, the \emph{observation model} (Observer) and the \emph{dynamics model} (Simulator); both commit a concrete claim before each action and either survive the next observation or get patched. Four further surfaces hold the action-side state used while the WM is still maturing: a \emph{strategy library} (IE), a \emph{probe context} (TE), \emph{failure routing} (Critic), and the \emph{goal and action-selection context} (TL). Table~\ref{tab:roles} lists each surface, its owner, and the artifacts written. Write $W^r_t$ for the slice of $W_t$ owned by role $r$. The surfaces interlock by design: each role's output is another role's input, so a single env step exercises the whole chain.

The WM and strategy library expose a small callable interface. The Observer parses $o_t$ into a per-step structured state $z_t$, which $\operatorname{render}(z_t)$ checks by round-tripping back to the screen. The Simulator commits to a one-step claim through $\hat z_{t+1} = \operatorname{predict}(z_t, h_{t+1}, a_t)$, where $h_t$ is an aggregated history carried across steps and updates as $h_{t+1} = \operatorname{history}(z_t, h_t, a_t)$. The strategy library exposes a harness-side $\operatorname{rollout}$ that generates imagined trajectories from $\operatorname{policy}(z_t, h_t)$. The asymmetry between $\operatorname{predict}/\operatorname{render}$ and the Observer's parse is deliberate: $\operatorname{render}$ and $\operatorname{predict}$ mirror functions that exist on the game-engine side and can therefore be committed as code in $\mathcal{P}_t$ and rerun by anyone, but parsing the grid $z_t \leftarrow o_t$ has no analogue engine function, so the Observer's task is transductive: each parse produces a per-step output, and no callable form is committed for reuse.

Carrying $h_t$ separately gives the agent traction on partially observable games: $z_t$ alone is insufficient when the current grid hides part of the state (occlusion, off-screen entities, latent timers), and $h_t$ is the durable summary of past observations $\operatorname{predict}$ reads alongside $z_t$. Full signatures, regression checks, and the role-context fields that surround these calls are deferred to Appendix~\ref{sec:agentic-workspace-details}.

\begin{table}[h]
\centering
\caption{\textsc{DreamTeam} roles, artifacts owned, and team feedback edges. The strategy library holds sub-goals $\mathcal{G}_t$ and policies $\Pi_t$ used before the dynamics model is reliable; the probe context records information-seeking actions for unfit regions; failure routing turns contradicted predictions into addressable patches; the goal/action context chooses plan, probe, or repair. \emph{Inductive output} is a callable committed to $\mathcal{P}_t$; \emph{transductive output} is a per-step production.}
\label{tab:roles}
\small
\setlength{\tabcolsep}{3pt}
\begin{tabular}{lllll}
\toprule
\textbf{Surface} & \textbf{Owner} & \textbf{Inductive output} & \textbf{Transductive output} & \textbf{Feedback to} \\
\midrule
Observation model & Obs & $\operatorname{render}()$ & $z_t$ & Sim \\
Dynamics model & Sim & $\operatorname{predict}()$, $\operatorname{history}()$ & $h_t$ & Obs \\
Strategy library & IE & $\mathcal{G}_t$, $\Pi_t$ & --- & --- \\
Probe context & TE & --- & action-sequence probes & --- \\
Failure routing & Crt & --- & feedback directives & IE, TE, TL \\
Goal/action context & TL & --- & chosen action sequence or policy & --- \\
\bottomrule
\end{tabular}
\end{table}
\vspace{-0.3cm}
\subsection{The WM Refinement Loop}
\label{sec:step-loop}

\textsc{DreamTeam}'s WM refines through a commit-then-retrodict loop. The Simulator's dynamics program is fixed in $W_t$ when action $a_t$ is committed; after the next observation $o_{t+1}$ arrives, the~observer encodes $z_{t+1}$ and the harness compares the prediction\footnote{The rerun loads $\operatorname{predict}$ from the live dynamics module, so a Simulator patch made during step $t$ is the version exercised both for this diff and for the regression replay below.} $\hat{z}_{t+1}=\operatorname{predict}(z_t, h_{t+1}, a_t)$ with $z_{t+1}$:
\[
e_t = \operatorname{diff}(\hat z_{t+1}, z_{t+1}).
\]
The harness diff is one source of refinement signal. The other is \emph{peer feedback}: roles audit each other's artifacts and route findings inline as $F_{a \to b}$, where role $a$ writes a directed comment to role $b$ about $b$'s artifact. Feedback carries information the diff does not: the audit reads reasoning chains and ontology choices in addition to value mismatches. Inside the WM the Observer audits the Simulator's $\operatorname{predict}$ and the Simulator audits the Observer's $z_t$ parse, so each maintainer's evaluation signal $e^r_t$ bundles the diff with the incoming peer feedback, matching the $\operatorname{Update}(W_t, e_t)$ form of Section~\ref{sec:workspace-optimization}:
\[
e^{\mathrm{sim}}_t = \big(e_t,\ F_{\mathrm{obs} \to \mathrm{sim}}\big),
\qquad
W^{\mathrm{sim}}_{t+1} \;=\; \operatorname{Update}\!\big(W^{\mathrm{sim}}_t,\ e^{\mathrm{sim}}_t\big),
\]
and symmetrically for the Observer with $e^{\mathrm{obs}}_t = (e_t,\ F_{\mathrm{sim} \to \mathrm{obs}})$.

In the \texttt{tu93} failure shown in Figure~\ref{fig:dreamteam-step}, $e_t$ is non-empty. The harness decomposes it into named~components and surfaces each to its artifact's owner: the Observer sees its per-field encoding mismatches, the Simulator sees its prediction errors. Here the diff has an observation component (the Observer had not named the obstacle sprite) and a dynamics component (the Simulator did not yet model contact as lethal); the resulting repairs add an \texttt{obstacle} entity and a lethal-contact rule. The regression checks report which previously-held cases the new program would break, informing the next round of edits. 

\subsection{From Probing to Planning}
\label{sec:probe-to-plan}

\begin{figure}[h]
\centering
\includegraphics[width=0.8\linewidth]{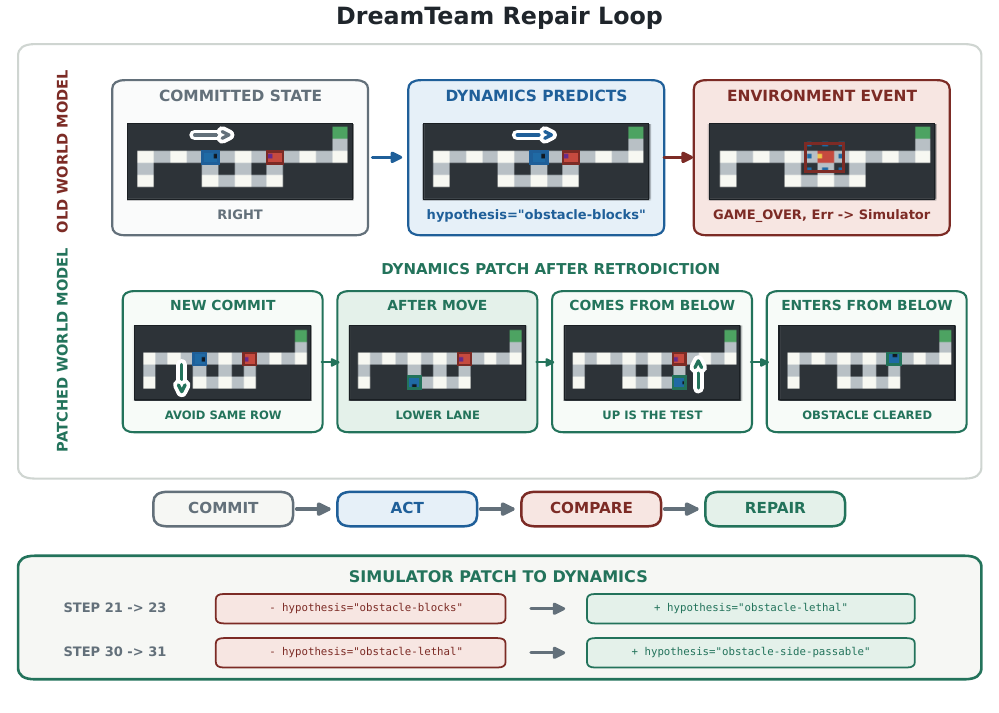}
\caption{A commit-then-retrodict repair on a real failure (game
\texttt{tu93}, level 1, step 22). The old WM commits
\texttt{RIGHT} on the current board and predicts that the obstacle will
block the player, but the game reaches \texttt{GAME\_OVER}.
The patched WM routes below the obstacle, arrives directly under
it, and enters from below, avoiding the lethal same-row
approach. The strip below the grid rows marks the loop, commit, act,
compare, repair. The compact Simulator patch shows the hypothesis
refinements that first make the obstacle lethal, then add a side-entry
hypothesis.}
\label{fig:dreamteam-step}
\vspace{-0.3cm}
\end{figure}

At each step, the three actor roles (TL, IE, TE) produce candidate actions from different signals. The IE takes the current WM state $(z_t, h_t)$, runs $\operatorname{rollout}$ for each policy $\pi \in \Pi_t$, and scores the resulting $(\hat z, \hat h)$ trajectories against the active sub-goal set $\mathcal{G}_t$; policies whose rollouts reach a sub-goal become candidate plans. The TE reads the recent step logs and the harness's retrodiction routing and proposes information-seeking action sequences for regions the WM has not yet fit. The TL sits at the sink: it inspects the IE-policy candidates and the TE-sequence candidates, judges whether the WM is reliable enough for rollout-based planning, and then commits an action or a policy as $a_t$.

Each actor updates from two error signals at the close of the step. First, the Critic audits every actor's proposals against the team's evidence and emits $F_{\mathrm{crt} \to \mathrm{TE}}$, $F_{\mathrm{crt} \to \mathrm{IE}}$, and $F_{\mathrm{crt} \to \mathrm{TL}}$, flagging assumption failures, ontology drift, or grinding loops. Second, the actor observes its own commitment fail: an IE policy whose rollout predicted progress and instead lost a level supplies its own counterexample, and a TE probe whose hypothesised effect did not appear supplies one too. Both inputs enter the actor's update via a bundled evaluation signal:
\[
e^{s}_t = \big(\delta^{s}_t,\ F_{\mathrm{crt} \to s}\big),
\qquad
W^{s}_{t+1} \;=\; \operatorname{Update}\!\big(W^{s}_t,\ e^{s}_t\big),
\qquad s \in \{\mathrm{TL}, \mathrm{IE}, \mathrm{TE}\},
\]
where $\delta^{s}_t$ is the actor's own hypothesis-vs-outcome diff, the actor-side analogue of the harness retrodiction diff that updates the WM surfaces.

The same workspace looks different at different points in a level. Early in level 1 of \texttt{tu93} (Figure~\ref{fig:dreamteam-step}), before the WM captured the obstacle's contact rules, actions such as touching a tile or walking into a wall served as probes: the TL had no reliable predictions to plan against. Once the surfaces stabilized and the strategy library held a tactic that fit, the TL committed to model-based plans. The first death triggered a repair that promoted obstacle contact from a blocking rule to a lethal one; for the next several steps the TL favored probes until the patch settled. The result is a control loop in which each action has an epistemic role as well as a game objective: it may confirm that the WM still fits, expose a counterexample for repair, or supply the evidence that localizes the next repair.            % DreamTeam
\section{Experiments}
\label{sec:experiments}

\begin{figure}[t]
\centering
\includegraphics[width=\linewidth]{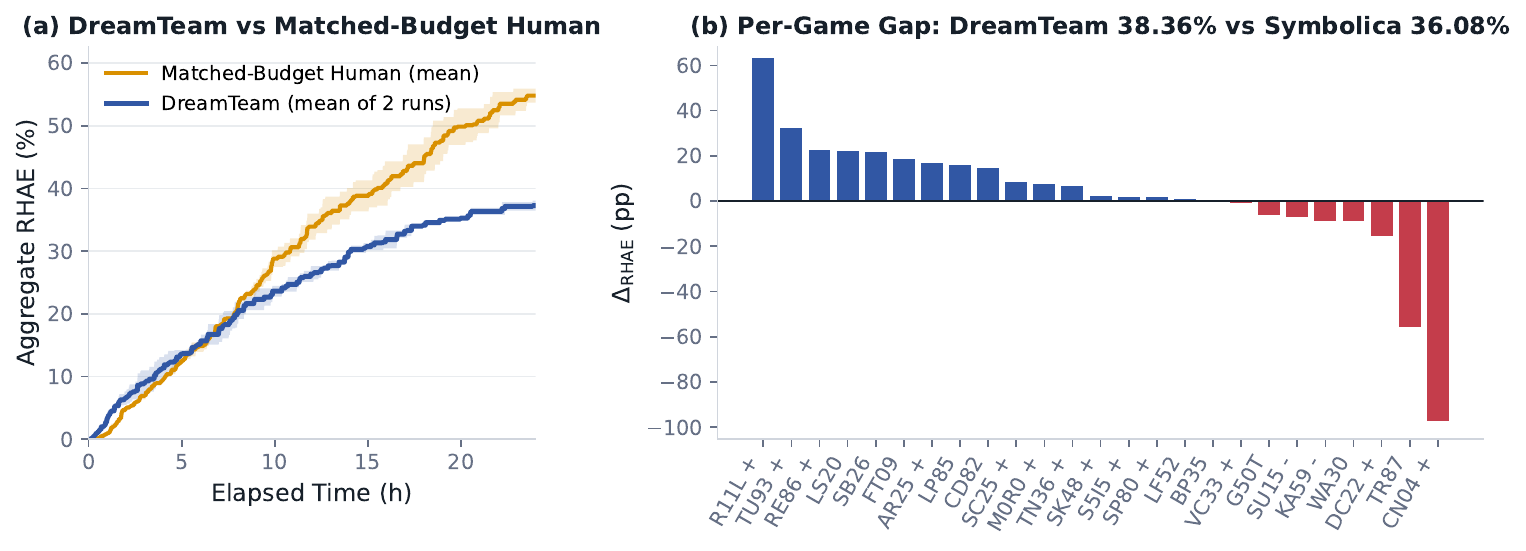}
\caption{\textbf{(a)} 24h matched-budget RHAE timeline averaged over our two runs (per-run range shaded): \textsc{DreamTeam} (blue) vs the matched-budget human (amber). \textbf{(b)} Per-game RHAE $\Delta_g$ vs Symbolica's single published run; each bar is the per-game mean of our two runs. $+$/$-$ next to a label marks a baseline change in human actions (15 games have changed, see Table~\ref{tab:benchmark-version-shift}).\vspace{-0.3cm}}
\label{fig:exp-headline}
\end{figure}

We evaluate \textsc{DreamTeam} on the 25-game ARC-AGI-3 public set
with the official RHAE scorer. See the ARC Prize scorecard record
for one of our two runs:
\url{https://arcprize.org/scorecards/831c83cf-b969-45fc-a6ce-27f9b3c4105c}. Figure~\ref{fig:exp-headline}(a)
plots aggregate RHAE sampled every minute against a matched-budget
human reference (the levels a baseline-efficient human would clear at
our action count). Averaged over our two runs, \textsc{DreamTeam} stays
ahead for the first $\sim\!5$~h (peak lead $\sim\!+4.5$~pp near hour~3);
the curves cross at $\sim\!5.3$~h, after which the human pulls
ahead through the 24-hour mark. The two runs sit in a tight
band around the mean. We attribute the late-game gap to the human's
stronger cross-level transfer: a mechanic understood at level~$\ell$
is re-used nearly cost-free at level~$\ell{+}1$, whereas the team
rebuilds parts of its workspace at each transition.

Figure~\ref{fig:exp-headline}(b) compares per-game RHAE against
Symbolica's Agentica SDK~\citep{arcgentica} on the same 25 games.
Averaged over our two runs, \textsc{DreamTeam} reaches mean RHAE
$\mathbf{38.36\%}$ vs Symbolica's single-run $36.08\%$. Each level
\textsc{DreamTeam} clears sits closer to the per-level cap, which the
triangular weighting amplifies. On the ten games whose version did
not change in the April~14 update the gap is $+3.42$~pp in our favor.

\textsc{DreamTeam}'s RHAE gain over Symbolica comes with
substantially fewer environment actions. Across the
25-game set and averaged over our two runs, \textsc{DreamTeam}
consumes a mean of $\mathbf{444}$ environment actions per game
against Symbolica's single published run of $\mathbf{16{,}081}$,
a $\mathbf{31\%}$ reduction in mean per-game actions. The largest
gains come from games where the baseline spends many actions on
a single level (whether the level is eventually cleared or not):
showcasing the advantage in planning over a learned world model. Restricted to successful
levels and pooled across our two runs, \textsc{DreamTeam} averages
$\mathbf{54.1}$ actions per solved level
vs Symbolica's $\mathbf{100.1}$, a $\mathbf{46\%}$ reduction; on
the 85 level indices that all of run~1, run~2, and Symbolica
cleared, the per-level counts are $\mathbf{49.0}$ (DT mean of two
runs) vs $\mathbf{57.3}$ ($-14.5\%$). Table~\ref{tab:exp-per-role}
reports per-role activity over the $22{,}180$ steps pooled across
our two runs; see full details in
Appendix~\ref{app:additional-experiments}.

\begin{table}[t]
\centering
\small
\caption{Per-role activity per env\_step, averaged over all
$22{,}180$ env\_steps across $25$ games and our two independent
runs. \emph{LLM} = LLM calls; \emph{In/Out~K} = input/output
tokens (thousands); \emph{Fb~s/r} = peer $@$-mentions
sent$/$received per step; \emph{Gen~s} = LLM generation seconds;
\emph{Code} = REPL blocks executed. \emph{TOTAL} sums per-role
columns; \emph{Gen~s} on the TOTAL row is the wall-clock seconds
per step (not the sum of per-role generations, since roles run in
parallel).}
\label{tab:exp-per-role}
\setlength{\tabcolsep}{4pt}
\begin{tabular}{l l rrr c rr}
\toprule
\textbf{Role} & \textbf{Model} &
\textbf{LLM} & \textbf{In~K} & \textbf{Out~K} &
\textbf{Fb~s/r} & \textbf{Gen~s} & \textbf{Code} \\
\midrule
Observer              & Opus 4.6 (none) & $1.97$ & $133.1$ & $3.6$ & $\;1.1/2.9\;$ & $\;68.5$  & $5.20$ \\
Simulator             & Opus 4.6 (none) & $1.93$ & $\;98.8$ & $4.0$ & $\;1.0/5.9\;$ & $\;82.9$  & $5.82$ \\
Inductive Explorer    & GPT-5.5 (high)  & $1.92$ & $140.2$ & $5.1$ & $\;3.2/8.7\;$ & $\;81.7$  & $3.17$ \\
Transductive Explorer & Opus 4.6 (low)  & $1.75$ & $\;75.5$ & $4.5$ & $\;0.9/5.7\;$ & $147.8$ & $2.83$ \\
Critic                & GPT-5.5 (high)  & $1.95$ & $154.1$ & $4.3$ & $22.8/2.0$    & $\;76.8$  & $2.83$ \\
Team Leader           & Opus 4.6 (low)  & $2.01$ & $\;91.3$ & $4.5$ & $\;5.3/9.3\;$ & $159.2$ & $2.49$ \\
\midrule
\textbf{TOTAL}        & n/a             & $\mathbf{11.5}$ & $\mathbf{692.9}$ & $\mathbf{26.0}$ & $\mathbf{34.4/34.4}$ & $\mathbf{205.2}$ & $\mathbf{22.3}$ \\
\bottomrule
\end{tabular}
\end{table}

\section{Limitations and Conclusion}
\label{sec:limitations-conclusion}

Our reported scores are bounded by time and cost budgets rather
than by system intelligence: the wall-clock cap sets the ceiling
for many games we did not finish, and a larger budget would
plausibly lift the headline numbers further. A second
infrastructure limitation is our reliance on hosted provider APIs,
whose latency and availability can drift over the course of a run.
Overloaded endpoints surface as timeouts and degraded per-step
throughput; although these effects are external to the agent
design they still materially affect any single-run number.

These are operational caveats. The conceptual claim of the paper
concerns what becomes trainable once weights are off the table,
and \textsc{DreamTeam} makes that state explicit and executable:
every prediction is a written commitment, every contradicted
commitment becomes a typed repair, and every repair is replayed
against earlier evidence before it is accepted. Treating the
workspace as the object of training opens a research agenda
parallel to the one that has organized weight-space learning for
decades, with its own questions about which artifacts to expose,
which counterexamples to retain, and which repair operators
generalize across environments. The present results are a first
step along that agenda, and we expect the methodology to extend
to other interactive settings where the model is fixed and
learning has to happen through the workspace around it.

% % ── Acknowledgments (hidden in submission mode) ──────────────────────
% \begin{ack}
% TODO: Acknowledgments and funding disclosure.
% \end{ack}

% ── References (do not count toward 9-page limit) ────────────────────
\bibliographystyle{plainnat}
\bibliography{references}

% ── Appendices (do not count toward 9-page limit) ────────────────────
\appendix
% ══════════════════════════════════════════════════════════════════════
% APPENDIX A: Extended Related Work
% ══════════════════════════════════════════════════════════════════════
\section{Extended Related Work}
\label{app:extended-related-work}

\paragraph{Model-based reinforcement learning and neural world models.}
Model-based RL learns predictive state and dynamics and then uses the resulting model for planning or policy improvement. World Models, Dreamer, and MuZero do this with neural latent states, learned transition models, and imagined rollouts~\citep{worldmodels2018,dreamerv2,dreamerv3,muzero}. The thing being trained, in all of these, is a parameter vector. \textsc{DreamTeam} keeps the computational shape (encode, maintain state, predict, reconstruct, roll out), but the substrate is code rather than weights, and feedback is a logged counterexample with a named owner rather than a gradient.

\paragraph{Generative interactive environments and playable world models.}
A recent line of generative-model work builds world models that are explicitly \emph{playable}: a user or policy issues actions and the model produces the resulting observations frame by frame in real time. Genie trains an action-conditioned video model that turns static internet videos into controllable interactive environments~\citep{bruce2024genie}. GameNGen demonstrates that a diffusion model can serve as a real-time game engine for DOOM, with neural rollouts standing in for the simulator~\citep{valevski2024gamengen}. DIAMOND uses diffusion world models on Atari and shows that visual fidelity inside the model materially affects what a policy trained inside it can learn~\citep{alonso2024diamond}. \textsc{DreamTeam}'s world model is playable in the same functional sense. The Simulator commits to one-step predictions through \texttt{predict}, the strategy library composes those predictions into multi-step \texttt{rollout}s, and the Inductive Explorer evaluates candidate sub-goals by playing them out inside the model before any real action is taken. The substrate differs from the generative-interactive-environments line: pixels rendered from a weight vector trained offline on large video corpora in their case, a small set of typed Python artifacts read, edited, and regression-tested online from at most dozens of interactions in ours. The shared design point is that the world model is the place where rollouts happen, and learning targets the fidelity of those rollouts to the real environment.

\paragraph{LLM agents with memory, tools, and code artifacts.}
LLM agents already adapt at inference time, through action traces, retrieved memories, reflections, tool calls, cognitive-architecture state, and persistent skill code~\citep{react,shinn2023reflexion,packer2023memgpt,sumers2023coala,voyager}. The writable state in most of this work is some combination of transcript, memory store, tool result, and skill library. The standard handling of a failure is to append another message, summary, reflection, or document to that pool. \textsc{DreamTeam} differs in where the failure goes: a misparse is sent to the observation model and not to anything else, a wrong action effect to the transition model, a tactic that overfits its own progress check to the strategy library, and a failure with no clean owner is converted into a probe. The point is the routing, not the size of the context. Meta-Harness~\citep{lee2026metaharness} optimizes the harness itself by searching over harness code across tasks via an agentic proposer that reads prior candidates' source, scores, and traces; workspace optimization runs the inner loop atop a fixed harness, adapting $W_t$ within a single episode.

\paragraph{Text-memory agents.}
A natural competitor class to typed-artifact workspaces is agents whose memory is mediated by natural-language text rather than typed, executable artifacts. Voyager grows a library of code skills retrieved by their natural-language descriptions, with routing done by description match rather than by artifact type~\citep{voyager}. Reflexion appends a verbal self-critique after each episode and reads it back as a textual memory that conditions the next attempt~\citep{shinn2023reflexion}; the unit of update is a completed trajectory, so feedback only becomes available between episodes. Generative Agents organize experience as a memory stream of natural-language observations scored by importance and recency at retrieval time~\citep{park2023generative}. MemGPT borrows operating-system metaphors to manage hierarchical text memory across a fast working context and a slower archival store~\citep{packer2023memgpt}. What unifies these systems is that memory is natural-language text retrieved at use time, where a passage is the unit of both write and read. \textsc{DreamTeam} instead writes into typed, executable artifacts: a contradicted commitment is routed to the observation, dynamics, or strategy slot that owns it, the patch is replayed against earlier evidence in a regression set, and credit assignment names a typed address rather than a passage to retrieve. The update cadence is also finer. Each environment action carries a one-step prediction that the next observation either confirms or contradicts, so a workspace edit is proposed within the trajectory itself, at the granularity of an individual transition. Adaptation therefore proceeds during the episode, and the regression set of earlier transitions surfaces consistency failures between candidate patches and prior evidence.

\paragraph{Programmatic context interaction and recursive inference.}
Recursive Language Models~\citep{zhang2025rlm} place a long prompt inside a Python REPL and let a root LLM examine it through code, optionally spawning recursive sub-calls on snippets and returning a value via \texttt{FINAL} or a REPL variable. The prompt becomes external state the model interrogates rather than tokens it must read. We share the move of pushing structure outside the model and letting code mediate between the model and a large external object. The settings differ in three ways. First, RLM operates within a single query and its REPL is ephemeral, while workspace optimization persists $W_t$ across steps and treats it as the trainable object. Second, RLM produces a transductive output (a value for the current query) and has no mechanism by which one call's failure localizes a repair that later calls will read; \textsc{DreamTeam} routes each contradicted commitment to a named owner and evaluates the patch against a regression set. Third, RLM decomposes one prompt with copies of the same model, while \textsc{DreamTeam} decomposes the agent itself into roles owning distinct surfaces (observation, dynamics, strategy, probing, critique, arbitration). RLM is a clean instance of a transductive-only design in our taxonomy of Section~\ref{sec:workspace-optimization}.

\paragraph{Program synthesis and executable world models.}
Program synthesis and library-learning systems treat programs as learned structure revised by examples, tests, or search~\citep{dreamcoder,grand2024lilo,romera2024funsearch}. Related LLM systems write planning domains, propose and refine world models from interaction, search over candidate hypotheses, or compile planning structure~\citep{guan2023llmp,hao2023rap,wong2023wordtoworld,webdreamer,worldllm,wang2024hypothesis}. On ARC-AGI-1/2, frozen-LLM agents evolve a candidate solution against demonstration grids, with Python programs on ARC-AGI-1 and natural-language instructions on ARC-AGI-2~\citep{berman2025arcagi2}. SOAR scales the same family by alternating evolutionary program search with hindsight LLM fine-tuning, reaching $52\%$ on the ARC-AGI public test~\citep{pourcel2025soar}; the inner search loop is similar to Berman's, but the proposer's weights are updated between trials, so adaptation is split between the candidate workspace and the parameter vector. Both designs run offline against demonstration sets, where the demonstration set serves as the regression evaluator and the candidate is the workspace. \textsc{DreamTeam} is closest to this line through its Simulator artifact, which evolves online against the regression set in the same way that offline candidate programs evolve against demonstration sets. The distinction is online control: each environment action produces a counterexample or confirmation, and a repair targets the current prediction while being evaluated against earlier traces. The executable world model is both the planning object and the object trained within the episode.

\paragraph{Online world modeling with curiosity-driven hypothesis revision.}
WorldLLM constructs a frozen-model world model online by maintaining a natural-language hypothesis about environment dynamics, refining it through Bayesian inference with a second LLM serving as the proposal distribution, and gathering evidence with an RL policy whose reward is the negative log-likelihood of observed transitions under the current predictor~\citep{worldllm}. The setting is close to \textsc{DreamTeam}: both keep the predicting LLM frozen, both treat the world model as the primary trainable surface, and both use a secondary LLM to propose updates from accumulated evidence. The two systems differ along three axes. First, the substrate. WorldLLM stores the world model as a single passage of natural-language hypothesis read by the predictor at inference time. \textsc{DreamTeam} stores it as executable Python with explicit \texttt{predict} and \texttt{rollout} entry points, so multi-step rollouts are evaluated by the runtime, and the workspace is partitioned into typed slots for observation, dynamics, strategy, probing, critique, and arbitration. Second, the verification of an update. WorldLLM accepts a revised hypothesis when it improves likelihood under the current evidence buffer. \textsc{DreamTeam} replays a loaded patch against prior transitions retained as a regression set and reports any consistency failures as counterexamples, so the check is explicit and falsifiable even though replay is feedback rather than an automatic rollback gate. Third, the cadence. WorldLLM alternates evidence collection with hypothesis revision in trials. \textsc{DreamTeam} proposes a patch after every action and treats the next observation as a one-step counterexample or confirmation. In our taxonomy, WorldLLM is a single-slot instance of workspace optimization in which the proposer is Bayesian, the verifier is likelihood, and the cadence is trial-level.

\paragraph{Iterative self-correction with execution feedback.}
A line of LLM-agent work treats outputs as drafts that the model revises against feedback from a critic or an executor. Self-Refine prompts the same model to critique and rewrite its own answer over multiple iterations~\citep{madaan2023selfrefine}, and Self-Debug pairs code generation with execution traces so that the model can localize and repair the offending lines~\citep{chen2023selfdebug}. These systems share with \textsc{DreamTeam} the move of routing a counterexample back into the artifact that produced it. \textsc{DreamTeam} differs in two structural respects. First, the artifact under repair is one of a small set of typed surfaces (observation, dynamics, strategy, probing) with bounded scope, so a contradicted commitment names a specific owner rather than revising a single prompt or program in place. Second, each loaded patch is replayed against earlier transitions retained as regression tests, so any disagreement with prior evidence is surfaced as explicit feedback alongside the current failure.

\paragraph{Test-time training and online adaptation under fixed parameters.}
Test-time training adapts a model on the test instance itself by running self-supervised updates before producing a prediction~\citep{sun2020ttt}. The strategy has been unusually effective on abstract-reasoning benchmarks. Fine-tuning on synthetic transformations of each ARC task before solving it raises accuracy substantially~\citep{akyurek2024ttt}, and nearest-neighbor variants improve language-model reasoning more broadly~\citep{hardt2024ttt}. More recent variants pursue the same goal under frozen weights: ArcMemo distills concept-level abstractions from prior solution traces into a lifelong memory retrieved at test time, reporting consistent gains on ARC-AGI without parameter updates~\citep{ho2025arcmemo}. These methods address the same problem \textsc{DreamTeam} targets: how a model adapts to an unfamiliar instance in the absence of offline data or task-specific rewards. The mechanism differs along two axes. First, gradient-based test-time training writes the update into the parameter vector, which requires weight access and discards the adaptation once the task ends, while memory-retrieval variants grow a flat pool of past abstractions accessed by similarity to the current input. \textsc{DreamTeam} instead writes the update into a typed workspace whose slots persist as inspectable text and code and survive deployment behind a fixed inference API. Second, the test-time-training update is driven by a self-supervised auxiliary objective derived from the test input, and the memory-retrieval update is driven by a similarity score, while the \textsc{DreamTeam} update is driven by a contradicted commitment routed to the typed artifact responsible and scored against a regression set, so credit assignment names a structural owner instead of producing a global parameter direction or a retrieval ranking.

\paragraph{Multi-agent LLM systems and computational graphs.}
Multi-agent LLM systems assign roles, exchange messages, decompose tasks, or optimize communication graphs~\citep{metagpt,zhuge2024gptswarm,agenticsurvey}. Conductor~\citep{nielsen2025conductor} learns an orchestrator's weights via reinforcement learning over worker topologies and per-worker prompts, while \textsc{DreamTeam} keeps every model frozen and locates learning in the routed patch and the regression set. Recent work formalizes the inverse direction, automated failure attribution: given a multi-agent failure, identify which agent and which step caused it~\citep{zhang2025attribution}. Current methods on the Who\&When benchmark reach only $53.5\%$ agent-level accuracy, which suggests that attribution is hard when agents communicate over unstructured natural-language messages and any agent could in principle be responsible for any commitment. \textsc{DreamTeam} sidesteps the inverse problem by construction: every prediction is emitted by exactly one typed slot, so the responsible owner is named at write time and need not be reconstructed afterwards. \textsc{DreamTeam} also uses roles, but as nodes in a workspace training graph. Each node owns part of the trainable state or update process: observation, dynamics, strategy, probing, critique, or arbitration. A prediction failure is not debate material. It is routed to an owner with bounded patch scope, and the regression set is replayed after the edit, with deltas surfaced as feedback for subsequent revisions.

\paragraph{ARC and interactive reasoning benchmarks.}
ARC-style benchmarks test abstraction and rule discovery under weak priors~\citep{chollet2019}. ARC-AGI-3 extends this pressure to interaction, where the task is hidden in the game and every exploratory action consumes budget~\citep{arcagi3}. Game-based agent benchmarks also stress planning, observation, and tool use~\citep{balrog,orak}. \textsc{DreamTeam} targets the narrower bottleneck of within-episode construction of a testable world model.

The released ARC-AGI-3 benchmark reports frontier-LLM baselines below one percent of human RHAE: Gemini~3.1 Pro at $0.37\%$, GPT-5.4 at $0.26\%$, Opus~4.6 at $0.25\%$, and Grok~4.20 at $0\%$~\citep{arcagi3}. A subsequent 160-replay analysis of GPT-5.5 and Opus~4.7 traces these scores to a missing structured world model: agents register local action effects without composing them into a stable theory of game dynamics, with Opus over-committing to a wrong theory and GPT-5.5 failing to compress evidence into a theory at all~\citep{arcprize_gpt55_opus47}. The strongest published agent on the released benchmark is Symbolica's Agentica SDK, an Opus~4.6 multi-role harness that reaches $36.08\%$ RHAE on $113$ of $182$ levels at approximately \$1{,}005~\citep{arcgentica}, while \textsc{DreamTeam} reaches $38.06\%$ on the post-update protocol described in Section~\ref{sec:experiments}. The earlier ARC-AGI-3 Agent Preview Competition produced complementary design points, including a non-LLM reinforcement-learning baseline~\citep{stochasticgoose} and a graph-based exploration agent~\citep{graphexplore}, both reported on the Preview environment and on game versions that predate the April~14 update~\citep{arcprize_changelog}, placing them outside the released-benchmark protocol used in our results.

\section{Algorithm and Pseudocode}
\label{app:algorithm}

This appendix expands the operational detail of the per-step loop summarized in Algorithm~\ref{alg:dreamteam-step}. Section~\ref{app:alg:step} traces the full one-step sequence; Section~\ref{app:alg:reflection} describes the reset and level reflection flows that wrap the step loop; Section~\ref{app:alg:scheduling} describes role scheduling and credit assignment by ownership; Section~\ref{app:alg:repair} specifies validation, regression, and the behavior of a failed repair.

\paragraph{Notation and concrete defaults.}
Throughout, we use \(W_t = (\mathcal{P}_t, \mathcal{L}_t)\) for the workspace at step \(t\), with \(\mathcal{P}_t\) holding its three executable artifacts (\texttt{observable.py}, \texttt{dynamics.py}, \texttt{strategy.py}) and \(\mathcal{L}_t\) holding the per-role Markdown logs (step log and level log). Alongside the workspace, the harness maintains the observation dataset \(\mathcal{D}_t\) as JSONL traces (\(z_{\text{encoded}}\), \(h_{\text{encoded}}\), \texttt{level\_constants}), and \(\mathcal{R}_t\) is the regression set drawn from \(\mathcal{D}_t\). Per-step quantities \(z_t\), \(h_t\), \(a_t\), and \(c_\ell\) refer to the parsed scene, the dynamics hidden state available at observation \(o_t\), the committed action, and the level constants. Thus the transition ending at \(o_t\) is evaluated as \(\operatorname{predict}(z_{t-1}, h_t, a_{t-1}, \ldots)\), with \(h_t\) produced by \(\operatorname{history}(h_{t-1}, z_{t-1}, a_{t-1}, \ldots)\). We use \(K\) for the catch-up bound on \(h\)-replay, \(d\) for the per-role round depth, and \(B\) for the per-step token budget. Unless overridden by the run configuration, runtime defaults are: round depth \(d = 4\) per role, hidden-state catch-up bound \(K = 20\) steps, RESET cooldown \(5\) actions, error-ledger capacity \(20\) entries per source per role, step wall-clock timeout \(360\) seconds, per-step token budget \(B = 250{,}000\) tokens across all roles, step-log retention of the last \(24\) appended sections, and a per-game cost cap of \(\$2{,}000\) (USD) above which the run terminates. Per-role launch delay defaults to one step for the Critic and Inductive Explorer and to zero for the other four roles.

\begin{algorithm}[h!]
\caption{One \textsc{DreamTeam} timestep}
\label{alg:dreamteam-step}
\begin{algorithmic}[1]
\Require New $o_t$ and a Workspace $W_t$ with logs $\mathcal{L}_t$ and the programs $\mathcal{P}_t$ in Table~\ref{tab:roles}
\Ensure Action sequence $\{a_i\}$ or $\operatorname{policy}_j()$
\State \textbf{Observer}: emit transient $z_t$; check $\operatorname{render}(z_t) \approx o_t$
\State \textbf{Simulator}: reshape $\operatorname{predict}$ and $\operatorname{history}$ so that $\operatorname{predict}(z_{t-1}, h_t, a_{t-1}) \approx z_t$
\State \textbf{TE}: analyze the grid, write hypotheses, and propose an information-seeking action sequence
\For{policy $\pi_i \in \Pi$}
    \State \textbf{IE}: $(\hat z, \hat h)^{\pi_i}_{t+1:t+k} \leftarrow \operatorname{rollout}(z_t, h_t, \pi_i)$; score against $\{\operatorname{sg}_j\}$
\EndFor
\State \textbf{Critic}: audits routed evidence and sends feedback $F_{\mathrm{Crt}\to s}$ to actors; each role updates its logs and programs
\State \textbf{TL}: select action sequence (from \textbf{TE}) or policy (from \textbf{IE})
\end{algorithmic}
\end{algorithm}

\subsection{One-step operational sequence}
\label{app:alg:step}

A timestep proceeds in ten phases. Phases (1) through (4) are harness-side preparation, phases (5) through (8) are role activity, and phases (9) through (10) commit the action and route the resulting evidence. The harness is a deterministic state machine; the only stochastic component within a step is the language-model call inside each role. Algorithm~\ref{alg:dreamteam-step-detail} states the same sequence as pseudocode; the prose below explains what each phase does and why.

\begin{algorithm}[t]
\caption{One \textsc{DreamTeam} timestep, expanded.}
\label{alg:dreamteam-step-detail}
\begin{algorithmic}[1]
\Require Workspace \(W_t = (\mathcal{P}_t, \mathcal{L}_t)\) and observation dataset \(\mathcal{D}_t\); environment with action set \(\mathcal{A}_t\); observation \(o_t\); committed previous action \(a_{t-1}\); roles \(\mathcal{R} = \{\text{Obs}, \text{Sim}, \text{IE}, \text{TE}, \text{Crit}, \text{TL}\}\); hyperparameters \((d, K, B)\).
\Ensure Workspace \(W_{t+1}\), action \(a_t\), transition event \(\tau_t\).
\Statex \textbf{Harness preparation.}
\State \(\text{ctx}_t \gets \textsc{BuildStepContext}(o_t, a_{t-1}, \text{level}_t, \text{metadata}_t)\)
\State \((h_t, k_t) \gets \textsc{Load}(h_t,\, K)\) \hfill \# direct hit \((h_t, 0)\) or replay \(k_t \in [1, K]\) steps
\If{\(t > 0\) \textbf{and} Observer and Simulator are present}
\State \(\hat z_t \gets \operatorname{predict}(z_{t-1}, h_t, a_{t-1}, c_\ell, m_t)\)
    \State \(e_{t-1} \gets \operatorname{diff}(\hat z_t, z_t)\); attach owner tag per error type
    \State Update per-hypothesis retrodiction accuracy in \texttt{HYPOTHESES}
\EndIf
\Statex \textbf{Role activity.}
\State Determine eager and deferred roles from start-after gates \(\mathcal{G}\)
\For{\(r \in \mathcal{R}\) eager}
    \State Build role message \(I_{r,t}\) from canonical block list; launch with stagger
\EndFor
\For{\(r \in \mathcal{R}\) deferred}
    \State Wait on \(\mathcal{G}_r\); rebuild \(I_{r,t}\) post-gate; launch
\EndFor
\For{each role \(r\)}
    \For{\(\rho \gets 1\) \textbf{to} \(d\)}
        \State Run REPL round; apply tool calls (read, write, append, edit)
        \For{each write to \(\mathcal{P}_t\)}
            \State \textsc{ParseValidate}; \textsc{ReloadModule}; \textsc{LiveRevalidate}
        \EndFor
        \If{output emitted \textbf{or} token cap \(B\) reached}
            \State \textbf{break}
        \EndIf
    \EndFor
\EndFor
\State Team Leader: wait gate \(\to\) read teammate logs \(\to\) action validation \(\to\) emit \(a_t\)
\If{\(a_t\) is invalid \textbf{or} no answer}
    \State \(a_t \gets \textsc{ChooseDefaultAction}(\text{preferred} = \texttt{None})\)
\EndIf
\Statex \textbf{Commit and route.}
\State \(o_{t+1}, r_{t+1}, \text{done}_{t+1}, \text{info}_{t+1} \gets \operatorname{env.step}(a_t)\)
\State Suppress diff if \(\text{info}_{t+1}\) reports a level transition
\State \(\tau_t \gets \textsc{BuildTransitionEvent}(a_t, o_t, o_{t+1}, \text{info}_{t+1})\); log to \(\mathcal{D}_t\)
\State \textsc{UpdateErrorLedger}: replay retrodiction across recent transitions; mark resolutions
\State Append step-log sections; trim to retention cap; save chats
\State \Return \(W_{t+1}, a_t, \tau_t\)
\end{algorithmic}
\end{algorithm}

\paragraph{Phase 1: Step context construction.}
The harness assembles a step context that captures the current grid \(o_t\), action history, level index, level-start step, and per-step metadata (available actions, action budget, diff pixel count from the previous transition). On a soft RESET, prior \(z_{\text{encoded}}\) entries for the current level are restored so that downstream readers bootstrap from level start rather than from an empty state.

\paragraph{Phase 2: Hidden state recovery.}
The dynamics hidden state \(h_t\) is loaded by an authoritative reader. A direct hit on \(h_{\text{encoded}}\) at step \(t\) returns \((h_t, 0)\). A miss triggers replay from the nearest anchor in the window \([t - K, t-1]\), with the catch-up bound set to \(K = 20\) steps by default. For each intermediate step \(s\), the harness computes
\[
h_s = \operatorname{history}(h_{s-1},\, z_{s-1},\, a_{s-1},\, c_\ell,\, m_s),
\]
where \(c_\ell\) are the level constants and \(m_s\) is the per-step metadata. The transition with \(a_{s-1} = \texttt{RESET}\) is handled by setting \(h_s \leftarrow \emptyset\), matching the convention used by the predict-side evaluator. Replay sources \(\operatorname{history}\) from the live \texttt{dynamics} module, so the most recent Simulator patch is the one actually exercised.

\paragraph{Phase 3: Live world-model evaluation.}
For \(t > 0\) and when both Observer and Simulator are present, the harness runs a one-step retrodiction on the previous transition. It loads \(z_{t-1}\), \(h_t\), and the action \(a_{t-1}\) that the Team Leader committed at the prior step, then computes
\[
\hat z_t = \operatorname{predict}(z_{t-1},\, h_t,\, a_{t-1},\, c_\ell,\, m_t)
\]
and the field-level diff \(e_{t-1} = \operatorname{diff}(\hat z_t,\, z_t)\). A second pass evaluates each named hypothesis in \(\texttt{HYPOTHESES}\) under the same input, recording per-hypothesis prediction accuracy that is later used by the rollout-readiness check. Any error raised during predict, history, or rollout is tagged with an owner hint: predict and history failures are routed to the Simulator, POLICIES and SUB\_GOALS failures to the Inductive Explorer, and render failures to the Observer.

\paragraph{Phase 4: Concurrent role launch with dependency ordering.}
All six roles are launched concurrently, but two start-after edges defer specific roles until upstream writes are visible:
\begin{itemize}
\item Simulator defers until Observer has appended its current-step \(z_{\text{encoded}}\).
\item Inductive Explorer defers until Observer has appended its current-step \(z_{\text{encoded}}\).
\end{itemize}
The dependency is expressed as a tool-gate of the form \texttt{append\_file:data/z\_encoded.jsonl:observer}, not as agent-completion. Tool-gate semantics are stronger than role-termination because a role can finish before its file write becomes visible to a downstream reader. Each role is also subject to a per-role launch delay (in steps) and a per-role stagger (in seconds) that smooths gateway load. Critic and Inductive Explorer use the default launch delay of one step, so Observer and Simulator have produced \(z_{\text{encoded}}\) and \(h_{\text{encoded}}\) before either evaluates the new scene. On step zero, a wait gate auto-skips required roles that are still delayed, so the Team Leader does not block on an absent Critic at game start.

\paragraph{Phase 5: Per-role context assembly.}
Each role's first-round message is assembled from a canonical, role-specific list of context blocks ordered as primacy, middle, recency. The primacy zone holds the step directive, level-transition status, round budget, and grid history. The middle zone holds analytical inputs (world-model status, retrodiction summary, render quality, metadata, teammate logs, asynchronous chat). The recency zone holds the role's own working materials (its executable artifact, its helper file, its level log, its step log). The call to action (workflow, checklist, output format) sits at the end so that the format constraint is closest to the model's emission point. The exact block order per role is fixed in the role-context registry.

\paragraph{Phase 6: Bounded-depth REPL rounds.}
Each role runs for at most \(d = 4\) rounds. A round consists of a model call, optional REPL execution, and tool calls (read, write, append, edit). Within a round the role can iterate freely over its REPL; between rounds, middleware emits structured update blocks reflecting any change to the live world-model surface. Rounds end when the role emits its required output (typically a step-log append plus, where applicable, a code edit) or when the depth bound is reached. A token budget ceiling of \(B = 250{,}000\) tokens applies across all roles in the step, and the default wall-clock timeout of \(360\) seconds applies to the entire step (the analyzed run in Appendix~\ref{app:additional-experiments} disables this cap); whichever fires first triggers a forced collection in which the harness uses whatever each role has produced.

\paragraph{Phase 7: Workspace patches with live validation.}
Patches to \(\mathcal{P}_t\) (\texttt{observable.py}, \texttt{dynamics.py}, \texttt{strategy.py}) are applied via \texttt{write\_file} or \texttt{edit\_file}. Three layers act on each patch:
\begin{enumerate}
\item Python source validation rejects writes that fail to parse and reports the syntax error back to the caller.
\item World-model loader middleware re-imports the patched module into \texttt{sys.modules} so subsequent harness operations see the new functions.
\item World-model validation middleware reruns predict and render on the latest transition under the new code and emits a delta block that names which test cases passed or failed. The delta is feedback rather than a hard gate, so the role can keep iterating within the round budget.
\end{enumerate}
Patches to text artifacts (step log, level log, helper) and JSONL data files (\(z_{\text{encoded}}\), \(h_{\text{encoded}}\), level constants) are appends rather than overwrites. The harness deduplicates JSONL writes that repeat the most recent record.

\paragraph{Phase 8: Team-leader arbitration.}
The Team Leader runs a wait gate that holds until the required upstream roles have either produced output or been excused by the delay logic. It then reads each role's current step-log section through a tail-follow channel, applies action validation (the action must be in the available-actions list and respect the RESET cooldown of \(5\) actions), applies policy-readiness validation (refuse to commit to a strategy rollout when the underlying world model has insufficient retrodiction support), and emits the action commitment in a structured output block. If the Team Leader fails to produce a valid answer within its safety round, the controller routes through the default-action policy specified in Section~\ref{app:alg:crash}, which is structured specifically to avoid implicit RESETs.

\paragraph{Phase 9: Environment step.}
The harness invokes \(\operatorname{env.step}(a_t)\) and receives \(o_{t+1}\), reward, done, and \(\operatorname{diff\_pixels}\). When the action triggers a level transition, the cross-level diff is suppressed (set to \texttt{None}) so that new-level pixels do not leak into the previous level's evidence. The harness logs a transition event with frame indices, frame decimation, and the final frame of the completed level when the SDK returns intermediate frames.

\paragraph{Phase 10: Counterexample routing and ledger update.}
The transition event is the canonical evidence for the next step. The error ledger middleware reruns retrodiction across the recent action history under the latest code and records, for each violation, a tuple \((\text{source},\, \text{step},\, \text{fields\_wrong},\, \text{action})\) where \texttt{source} is either \texttt{render} (Observer-owned) or \texttt{predict} (Simulator-owned). New entries are announced to the owning role at the next step through a single-shot WM Update block. Ledger entries that pass under the latest code are marked \texttt{resolved\_at} and the role sees a one-line resolution notice exactly once. The persistent ledger surfaces unresolved entries until they are repaired or trimmed by the per-source per-role cap of \(20\) entries. Step logs are likewise trimmed to the most recent \(24\) appended sections. Critic, Inductive Explorer, Transductive Explorer, and Team Leader read the ledger as analytical input but do not own its entries.

\subsection{Reflection and recovery flows}
\label{app:alg:reflection}

The per-step loop of Section~\ref{app:alg:step} runs unmodified for the common case in which the Team Leader commits to a non-RESET action and the environment returns a normal frame. Three flows wrap that loop and trigger when the harness detects a state that warrants reflection: reset reflection when a role proposes RESET, level reflection when a level completes, and crash recovery when an exception escapes a role's safety round. Each flow uses the same agent infrastructure as the step loop but with a different directive, a different context surface, and a different commit semantics.

\paragraph{Reset reflection.}
\label{app:alg:reset}
A RESET commitment is treated as a workspace event rather than a normal action. When the Team Leader proposes RESET in Phase 8, the harness intercepts the commitment before invoking \(\operatorname{env.step}(\texttt{RESET})\) and runs a two-phase reflection. In phase one, all six roles read a \emph{reset-reflection directive} that asks whether the proposal is justified, with a context surface that uses the recent action history of the current level rather than the new-level surface. The Team Leader emits one of three decisions: REJECT (no environment action; the proposal is withdrawn and play continues from the same scene), RESET (environment reset, workspace context preserved; we also call this a \emph{soft RESET}), or RESET\_FRESH (environment reset, strategy and observation context wiped). The same reflection runs unconditionally on a GAME\_OVER recovery, with \texttt{harness\_triggered = True} so that the REJECT option is masked and only the two real reset variants remain. A per-level cap on REJECT decisions blocks indefinite refusal: once the cap is reached on a level, REJECT is removed from the allowed set and the directive narrows to RESET versus RESET\_FRESH. After a RESET executes, a cooldown of \(5\) environment actions blocks any subsequent RESET commitment, with the gate enforced both at directive-render time and at the pre-\(\operatorname{env.step}\) assertion.

\paragraph{Level reflection.}
\label{app:alg:level}
When the environment reports \(\text{levels\_completed}_{t+1} > \text{levels\_completed}_t\), the harness pauses the step loop and runs a level-reflection round. All six roles activate concurrently, each receives a \emph{level-reflection directive} and a context surface restricted to the completed level (no new-level pixels, no new-level metadata), and writes a reflection section to its \texttt{*\_level\_log.md}. No environment action is consumed. The harness checks afterwards that each role appended a non-trivial section, logs a warning when a role wrote less than 80 bytes, and clears the activation segment so that the next-level step loop starts as a fresh segment. Level logs accumulate across levels and survive RESET; they constitute the long-term memory that conditions later levels of the same game.

\paragraph{Crash recovery.}
\label{app:alg:crash}
A crash is any unhandled exception that escapes a role's safety round. The recovery contract has four rules. First, the harness logs a Python traceback and emits a structured \texttt{crash\_recovered} event to every role's logger. Second, it picks a fallback action through a fixed priority chain:
\begin{enumerate}
\item the \emph{preferred} action when one is available from an in-flight multi-action batch (the next planned action that has not yet executed);
\item \texttt{UNDO} when \texttt{UNDO} is in the available-actions set;
\item the most recent non-RESET action committed during the current run;
\item a step-zero scaffold of \texttt{USE}, \texttt{CLICK 32 32}, or \texttt{UP}, whichever is first available.
\end{enumerate}
Third, the fallback is executed through the same single-action path as a normal commitment, so all bookkeeping (transition event, ledger update, step-log trim, agent-chat save) fires. Fourth, the harness sets a one-shot crash-recovery banner that is rendered at the top of the next step's primacy zone for every role, so the Team Leader sees what was tried and why on the next iteration. The chain refuses to emit RESET as a fallback under any circumstance; when no safe non-RESET option exists, the harness raises rather than silently committing to a state-erasing action.

\subsection{Role scheduling and credit assignment}
\label{app:alg:scheduling}

\paragraph{When each role acts.}
Six roles share the step. Each one has a fixed input surface and a fixed output surface defined by the file registry:
\begin{itemize}
\item Observer reads \(o_t\), prior \(z_{\text{encoded}}\), and reconstruction checks; writes \texttt{observable.py}, \(z_{\text{encoded}}\), \texttt{level\_constants}, and its step log.
\item Simulator reads \(z_{\text{encoded}}\) (via the start-after gate), action history, prior \(h_{\text{encoded}}\), and ledger entries with source \texttt{predict}; writes \texttt{dynamics.py} (which exposes \texttt{predict}, \texttt{history}, and \texttt{HYPOTHESES}), \(h_{\text{encoded}}\), and its step log.
\item Inductive Explorer reads \(z_{\text{encoded}}\) and \(h_{\text{encoded}}\) with diffs, retrodiction summaries, sub-goal acquisition state, and policy rollout results; writes \texttt{strategy.py} (which exposes \texttt{SUB\_GOALS} and \texttt{POLICIES}) and its step log.
\item Transductive Explorer reads grid history, the live \texttt{HYPOTHESES} table with per-hypothesis retrodiction accuracy, and teammate proposals; writes probe proposals, hypothesis tags, and its step log.
\item Critic reads the union of teammate step logs, the data-quality report, the goal-acquisition view, and policy rollouts; writes critique notes, owner assignments, and its step log.
\item Team Leader reads grid history, world-model status, goal acquisition, policy rollouts, teammate step logs, and asynchronous chat; writes the action commitment and its step log.
\end{itemize}

\paragraph{How handoffs occur.}
Handoffs are mediated by the shared file system and tool-gate dependencies. A role finishes when it writes its required outputs or exhausts its depth. Other roles read those outputs through tail-follow middleware that surfaces the most recent appended sections. Asynchronous chat is the one channel that does not flow through file appends; it carries short messages between roles within a step and is rendered at a fixed slot in each role's middle zone. The dependency graph is a directed acyclic graph rooted at the Observer and terminating at the Team Leader, with the Critic as a parallel evaluator that consumes from the four worker roles and is consumed by the Team Leader.

\paragraph{Credit assignment by ownership.}
A retrodiction failure carries an owner tag computed at evaluation time: \texttt{predict} failures go to the Simulator, \texttt{render} failures to the Observer, \texttt{POLICIES} or \texttt{SUB\_GOALS} failures to the Inductive Explorer. The owner tag is a string written into the ledger entry, not a gradient. The patch that follows is bounded to the artifact owned by that role (\texttt{observable.py} for the Observer, \texttt{dynamics.py} for the Simulator, \texttt{strategy.py} for the Inductive Explorer), so the address of the repair is structural rather than statistical. No model parameters change during a run. The persistent training signal is the workspace itself, and credit assignment is the routing of a counterexample to a typed slot.

\subsection{Repair feedback}
\label{app:alg:repair}

\paragraph{Validation checks.}
Three checks run in sequence whenever a role writes an executable artifact. First, a Python source validator parses the new file and reports the syntax error if parsing fails. Second, the world-model loader registers the new module in \texttt{sys.modules}, replacing the previous version. Third, world-model validation reruns predict and render on the live transition window and reports per-field accuracy. The first check rejects unparseable code outright. The second and third checks are observational: they record the post-edit state and emit a feedback block, but they do not roll back the file. The role observes the consequence of its edit on the same step.

\paragraph{Regression replay.}
After a patch lands, the error-ledger middleware runs a fresh retrodiction summary across the recent action history for the current level. Each prior transition is replayed as
\[
\hat z_s = \operatorname{predict}(z_{s-1},\, h_s,\, a_{s-1},\, c_\ell,\, m_s),
\]
and the diff against \(z_s\) is recorded per field. A ledger entry whose retrodiction now passes is marked \texttt{resolved\_at = t}, and the resolution notice is queued to the owning role exactly once. A ledger entry that still fails remains unresolved with its original owner. The role sees the entry on the next step in the persistent ledger block, with its accumulated diff fields rather than only the most recent failure.

\paragraph{Tie-breaking and conflict.}
Two situations require a deterministic decision. First, when a single transition has both a parse mismatch on \(z_{t-1}\) and a prediction mismatch on \(z_t\), the Observer is held responsible for the parse component and the Simulator for the prediction component, because the diff structure separates per-field encoding errors from per-field prediction errors. Second, when two roles edit overlapping responsibilities (for example, the Inductive Explorer adding a progress predicate that the Simulator has not modeled yet), the Critic flags the conflict in its step log and the Team Leader either delays the commitment for one step or commits to a probe rather than a planned action. The schema-level disjointness defined in Appendix~\ref{app:artifact-schemas} prevents the two roles from writing the same field through normal patches.

\paragraph{Behavior after a failed repair.}
A failed repair produces three observable consequences. The ledger entry stays unresolved, so the owner sees the same routed counterexample on the next step. The world-model status block reports the post-edit accuracy at the field level, so the owner sees which fields are still wrong. The Critic, reading the ledger as input, can flag the patch as overfitting if it resolves the latest failure but breaks earlier transitions; in that case the entry for an earlier step transitions from resolved to unresolved and the Critic's note carries the regression delta. There is no automatic rollback. Repeated failures lengthen the entry's footprint on the persistent ledger, which is the signal the Team Leader uses to commit to a probe rather than a planned action.

\paragraph{Alternative role decompositions.}
The six-role decomposition is one cut through the labor of constructing a testable world model. Two alternatives appear in the literature and clarify what the cut buys. A single-agent decomposition collapses \(\{\text{Obs}, \text{Sim}, \text{IE}, \text{TE}, \text{TL}, \text{Crit}\}\) into one function and folds the patch routing into a single inner monologue; the workspace is a single artifact and the Critic role disappears. A two-agent decomposition splits the team into a world modeler and a planner, leaving the rest implicit; this preserves the surface boundary between dynamics and strategy but loses ownership-level patch addresses for parsing, probing, and critique. The six-role cut chosen here has one role per typed surface, which is what makes the routing in Section~\ref{app:alg:step} address-stable: a contradicted commitment names a specific owner because each role owns exactly one executable artifact or one context surface.

\section{Artifact Schemas and Seed Workspace}
\label{app:artifact-schemas}

This appendix gives the schemas and seed-workspace details deferred from Sections~2 and~4. The workspace components and the observation dataset map to concrete storage: \(\mathcal{P}_t\) holds Python interfaces, \(\mathcal{L}_t\) holds Markdown role logs, and \(\mathcal{D}_t\) holds JSONL traces alongside the workspace; the harness draws \(\mathcal{R}_t\) from the JSONL trace. Section~\ref{sec:dreamteam} gives the role-by-role layout, and Appendix~\ref{app:execution-details} covers invocation order, context assembly, and runtime mechanics.

\subsection{Workspace schemas}
\label{sec:workspace-schemas}

The workspace is typed by four dataclasses. \texttt{ZState} stores the per-step structured observation, \texttt{HState} stores the dynamics hidden state propagated across steps, \texttt{LevelConstants} stores level-scoped invariants, and \texttt{Metadata} stores harness-provided context. The Observer owns \texttt{ZState} and \texttt{LevelConstants}; the Simulator owns \texttt{HState}; the harness writes \texttt{Metadata} and the agent reads it. Each is a flat dictionary at runtime, so the agent edits the dataclass declaration and the encoder writes a new JSON-compatible dictionary on the next step.

\paragraph{Observation state (\texttt{ZState}).}
\texttt{ZState} captures everything the Observer encodes from the current grid that may change between steps. The seed schema retains five fields and the Observer extends or prunes them as the game's vocabulary becomes clearer.

\begin{lstlisting}[basicstyle=\ttfamily\small, frame=single, xleftmargin=1em]
@dataclass
class ZState:
    object_positions: dict = field(default_factory=dict)
    object_states:    dict = field(default_factory=dict)
    sprite_overrides: dict = field(default_factory=dict)
    hud_values:       dict = field(default_factory=dict)
    event_objects:    dict | None = None
\end{lstlisting}

\texttt{object\_positions} maps an entity name to its instance positions \texttt{[[row, col], $\dots$]}. \texttt{object\_states} stores per-entity mutable values (toggles, counters, modes). \texttt{sprite\_overrides} records per-step deviations from \texttt{LevelConstants.objects}, keyed by \texttt{name} or \texttt{name:idx} with optional \texttt{rotation}, \texttt{pixels}, or \texttt{visible}. \texttt{hud\_values} maps each HUD indicator name to its numeric value. \texttt{event\_objects} encodes a transient mini-animation triggered by the action (\texttt{frame\_count} and a list of \texttt{regions} cross-referenced with entities), or \texttt{None} when no event occurred.

\paragraph{Dynamics hidden state (\texttt{HState}).}
\texttt{HState} stores only the recurrent state that \texttt{history()} needs to propagate forward. Long-lived dynamics knowledge that should survive a soft reset lives in code-level constants such as \texttt{LEARNED\_EFFECTS} rather than in \texttt{HState}.

\begin{lstlisting}[basicstyle=\ttfamily\small, frame=single, xleftmargin=1em]
@dataclass
class HState:
    object_velocities: dict = field(default_factory=dict)
    event_triggers:    dict = field(default_factory=dict)
    game_state:        dict = field(default_factory=dict)
\end{lstlisting}

\texttt{object\_velocities} maps an entity name to its step-wise velocity \texttt{[dr, dc]}. \texttt{event\_triggers} accumulates \texttt{\{count, total\_frames\}} per \texttt{"ACTION:entity"} pair. \texttt{game\_state} is a free-form dictionary that the Simulator widens as the game's state machine becomes clear (for instance \texttt{\{"gravity\_down": True, "selected\_piece": "piece\_0"\}}).

\paragraph{Level constants (\texttt{LevelConstants}).}
Per-level invariants live in \texttt{LevelConstants}. They are written to \texttt{data/level\_constants.jsonl} as an append-only stream so the harness can rerender past levels for retrodiction.

\begin{lstlisting}[basicstyle=\ttfamily\small, frame=single, xleftmargin=1em]
@dataclass
class LevelConstants:
    background_color: int  = 0
    camera:           dict = field(default_factory=dict)
    walls:            list = field(default_factory=list)
    objects:          dict = field(default_factory=dict)
    hud:              list = field(default_factory=list)
\end{lstlisting}

\texttt{background\_color} is the dominant background pixel. \texttt{camera} carries an optional viewport \texttt{\{vx, vy, vw, vh, padding\_color\}} for upscaled or letterboxed levels. \texttt{walls} is a static \texttt{64$\times$64} overlay with \texttt{-1} marking transparent cells. \texttt{objects} maps each entity name to either a 2D pixel array or a richer record \texttt{\{pixels, layer, rotation\}}. \texttt{hud} is a list of indicator templates such as horizontal or vertical bars.

\paragraph{Metadata (\texttt{Metadata}).}
\texttt{Metadata} is harness-owned and read-only for the agent. It is updated before each call to \texttt{history()}, \texttt{predict()}, or \texttt{policy()}.

\begin{lstlisting}[basicstyle=\ttfamily\small, frame=single, xleftmargin=1em]
@dataclass
class Metadata:
    step:              int  = 0
    level:             int  = 0
    available_actions: list = field(default_factory=list)
    grid_size:         int  = 64
\end{lstlisting}

\subsection{Function signatures}
\label{sec:function-signatures}

Three artifacts expose a small callable interface. The harness loads a fixed set of named exports from each artifact: \texttt{render} and \texttt{render\_event} from \texttt{observable.py}; \texttt{predict}, \texttt{history}, and \texttt{HYPOTHESES} from \texttt{dynamics.py}; \texttt{SUB\_GOALS} and \texttt{POLICIES} from \texttt{strategy.py}. Table~\ref{tab:harness-exports} summarizes the entry points.

\begin{table}[h]
\centering
\caption{Harness entry points loaded into the agent's REPL. Each artifact declares a fixed named export set; new helpers are private to the artifact.}
\label{tab:harness-exports}
\small
\setlength{\tabcolsep}{6pt}
\begin{tabular}{lll}
\toprule
\textbf{Artifact} & \textbf{Exports} & \textbf{Owner} \\
\midrule
\texttt{harness.py}    & \texttt{Metadata}                              & Harness \\
\texttt{observable.py} & \texttt{render}, \texttt{render\_event}        & Observer \\
\texttt{dynamics.py}   & \texttt{predict}, \texttt{history}, \texttt{HYPOTHESES} & Simulator \\
\texttt{strategy.py}   & \texttt{SUB\_GOALS}, \texttt{POLICIES}         & Inductive Explorer \\
\bottomrule
\end{tabular}
\end{table}

\paragraph{Observation model.}
\texttt{render} reconstructs a \texttt{64$\times$64} grid from the encoded state and the level invariants. \texttt{render\_event} reconstructs the sampled transient frames associated with an event, returning an empty list when no event occurred.

\begin{lstlisting}[basicstyle=\ttfamily\small, frame=single, xleftmargin=1em]
def render(z_encoded: dict, constants: dict) -> list[list[int]]:
    """Reconstruct a 64x64 grid from (z_encoded, constants).
    Layers (back to front): background, walls, entities (sorted by
    layer, applying overrides and rotations), upscaling, HUD."""

def render_event(z_encoded: dict, constants: dict) -> list[list[list[int]]]:
    """Reconstruct the sampled event frame sequence, or [] if no event."""
\end{lstlisting}

The pixel-accuracy loss compares \texttt{render(z\_encoded, constants)} against the observed grid; the event-accuracy loss compares \texttt{render\_event(z\_encoded, constants)} against the observed event frames. Both losses surface to the Observer as named retrodiction errors.

\paragraph{Dynamics model.}
\texttt{predict} returns the next-step \texttt{ZState}-shaped dictionary from the previous state, the recurrent \texttt{HState}, and the action just taken. \texttt{history} returns the next \texttt{HState} from the previous \texttt{HState}, the previous observation, and the action.

\begin{lstlisting}[basicstyle=\ttfamily\small, frame=single, xleftmargin=1em]
def predict(z_prev: dict, h_encoded: dict, action_prev: str,
            constants: dict, metadata: dict,
            hypothesis: str | None = None) -> dict:
    """Project z_predicted in ZState shape from (z_prev, h_encoded,
    action_prev). Branch on `hypothesis` for uncertain mechanics; when
    omitted, the highest-probability entry of HYPOTHESES is used."""

def history(h_prev: dict, z_prev: dict, action_prev: str,
            constants: dict, metadata: dict) -> dict:
    """Propagate HState forward from (h_prev, z_prev, action_prev)."""

HYPOTHESES: dict[str, float]  # name -> probability, sums to 1.0
\end{lstlisting}

\texttt{HYPOTHESES} is a probability mass over named hypotheses about uncertain mechanics. The harness retrodicts every entry and uses per-hypothesis residuals to update the mass; the Simulator branches on \texttt{hypothesis} inside \texttt{predict} only for the parts of the dynamics it cannot yet commit to deterministically.

\paragraph{Strategy library.}
The Inductive Explorer maintains two named export tables. Each entry is a state-dependent callable that reads the workspace state and returns either an action (for policies) or a progress tuple (for sub-goals).

\begin{lstlisting}[basicstyle=\ttfamily\small, frame=single, xleftmargin=1em]
def policy_fn(z_encoded: dict, h_encoded: dict,
              constants: dict, metadata: dict) -> str: ...

def sub_goal_fn(z_encoded: dict, h_encoded: dict,
                constants: dict, metadata: dict
               ) -> tuple[bool, float, str]:  # (achieved, progress, reason)
    ...

POLICIES:  dict[str, callable]  # name -> policy_fn
SUB_GOALS: dict[str, callable]  # name -> sub_goal_fn
\end{lstlisting}

Policies that ignore workspace state are logged as probes in the Transductive Explorer's step log rather than registered in \texttt{POLICIES}. The sub-goal triple returns whether the goal is satisfied, a progress fraction in $[0, 1]$, and a short \texttt{reason} string the Critic and Team Leader read.

\paragraph{Rollout primitive.}
The $\operatorname{rollout}$ used in Algorithm~\ref{alg:dreamteam-step} is a harness composition: given $(z_t, h_t)$ and a policy $\pi$, it iterates \(a_{t+i} = \pi(z_{t+i}, h_{t+i}, \dots)\), \(h_{t+i+1} = \texttt{history}(h_{t+i}, z_{t+i}, a_{t+i}, \dots)\), and \(\hat z_{t+i+1} = \texttt{predict}(z_{t+i}, h_{t+i+1}, a_{t+i}, \dots)\) for \(i < k\), scoring the trajectory against \texttt{SUB\_GOALS} at each step. Rollout-readiness is the conjunction of well-defined \texttt{predict}, \texttt{history}, and the relevant \texttt{POLICIES} and \texttt{SUB\_GOALS} entries.

\subsection{Seed workspace artifacts}
\label{sec:seed-artifacts}

When a game starts, the harness deploys a fixed set of files into the team's shared directory. Existing files are preserved across resets so that learned content survives. The deployment list is summarized in Table~\ref{tab:seed-files}.

\begin{table}[h]
\centering
\caption{Seed file inventory. The harness installs these files at game start; existing files are preserved across soft resets so that learned schemas, dynamics, and strategy survive.}
\label{tab:seed-files}
\small
\setlength{\tabcolsep}{6pt}
\begin{tabular}{lll}
\toprule
\textbf{Path} & \textbf{Purpose} & \textbf{Owner} \\
\midrule
\texttt{world\_model/harness.py}      & \texttt{Metadata} dataclass and shared utilities & Harness \\
\texttt{world\_model/observable.py}   & \texttt{ZState}, \texttt{LevelConstants}, \texttt{render}, \texttt{render\_event} & Observer \\
\texttt{world\_model/dynamics.py}     & \texttt{HState}, \texttt{predict}, \texttt{history}, \texttt{HYPOTHESES}, \texttt{LEARNED\_EFFECTS} & Simulator \\
\texttt{world\_model/strategy.py}     & \texttt{SUB\_GOALS}, \texttt{POLICIES} tables                 & Inductive Explorer \\
\texttt{data/z\_encoded.jsonl}        & Per-step \texttt{ZState} trace                  & Observer \\
\texttt{data/h\_encoded.jsonl}        & Per-step \texttt{HState} trace                  & Simulator \\
\texttt{data/level\_constants.jsonl}  & Append-only \texttt{LevelConstants} stream      & Observer \\
\texttt{*\_step\_log.md}              & Per-role per-step working log                   & Each role \\
\texttt{*\_level\_log.md}             & Per-role end-of-level summary                   & Each role \\
\texttt{*\_helper.py}                 & Per-role REPL helpers, auto-loaded each step    & Each role \\
\bottomrule
\end{tabular}
\end{table}

\paragraph{Seed variants.}
Two seed variants are available: a \emph{simple} seed and a \emph{rich} seed. The simple seed retains \texttt{ZState} fields \texttt{object\_positions}, \texttt{object\_states}, and \texttt{event\_objects}, and \texttt{LevelConstants} fields \texttt{background\_color}, \texttt{walls}, and \texttt{objects}. The rich seed adds \texttt{sprite\_overrides} and \texttt{hud\_values} to \texttt{ZState} and \texttt{camera} and \texttt{hud} to \texttt{LevelConstants}; its dynamics seed adds \texttt{game\_state} to \texttt{HState} and broader \texttt{LEARNED\_EFFECTS} fields including rotation deltas and visibility. Variant selection is per-game, controlled by the league configuration. The simple seed encourages a conservative encoding that the Observer extends only when forced by a parse failure; the rich seed biases toward upscaled and HUD-heavy games where overlays appear in the first few frames.

\paragraph{Seed defaults.}
The seeded \texttt{render} stamps the background color, applies the wall overlay where pixels are nonnegative, then stamps each entity's sprite at every position from \texttt{object\_positions}, with \texttt{sprite\_overrides} replacing the base pixels when present. The seeded \texttt{predict} carries each entity forward by the action's effect from \texttt{LEARNED\_EFFECTS} when known, otherwise by the entity's velocity from \texttt{HState} when known, otherwise leaves the position unchanged. The seeded \texttt{history} carries forward observed velocities and accumulates event-trigger statistics from \texttt{z\_prev["event\_objects"]}. The seeded \texttt{HYPOTHESES} contains a single primary entry; the Simulator widens it once retrodiction surfaces a mechanic it cannot commit to deterministically. The seeded \texttt{SUB\_GOALS} and \texttt{POLICIES} tables are empty and the Inductive Explorer fills them as rollout-supported plans accumulate.

\paragraph{Code-level constants.}
The dynamics artifact also exposes \texttt{LEARNED\_EFFECTS}, a code-level dictionary that stores confirmed action-to-entity effects discovered through retrodiction. Its schema is

\begin{lstlisting}[basicstyle=\ttfamily\small, frame=single, xleftmargin=1em]
LEARNED_EFFECTS: dict[str, dict[str, dict]]
# {ACTION: {entity_name: {dr: int, dc: int, drot: int, visible: bool}}}
\end{lstlisting}

where the inner record carries the row delta, column delta, rotation delta, and visibility change applied to each instance of \texttt{entity\_name} when the player commits \texttt{ACTION}. Storage at the code level (rather than inside \texttt{HState}) is what lets a confirmed effect survive a soft reset that wipes the runtime trace.

\paragraph{Helper files.}
Each role has a persistent \texttt{*\_helper.py} file that the harness auto-loads into the role's REPL at the start of every step. Helpers are private to a role and do not export through \texttt{HARNESS\_FUNCTIONS}; they exist so a role can build internal utilities (entity-matching predicates, rollout scoring helpers, debugging probes) without cluttering the four exported artifacts. Helpers are part of the workspace in the sense of Section~\ref{sec:dreamteam}: edits accumulate across steps, the Critic can read them, and they are subject to the same accept-on-evidence discipline as the executable artifacts.

\subsection{Schema disjointness}
\label{sec:schema-disjointness}

The workspace is partitioned so that each piece of evidence is owned by exactly one artifact. Disjointness gives the harness a deterministic routing rule: a counterexample component lands at the artifact whose schema covers the violated field. Table~\ref{tab:field-ownership} lists each field and its owner.

\begin{table}[h]
\centering
\caption{Field ownership across the workspace. Each field appears in exactly one schema and is written by exactly one role; retrodiction errors over a field are routed to that role.}
\label{tab:field-ownership}
\small
\setlength{\tabcolsep}{4pt}
\begin{tabular}{llll}
\toprule
\textbf{Schema} & \textbf{Field} & \textbf{Owner} & \textbf{Repair signal} \\
\midrule
\texttt{ZState}          & \texttt{object\_positions}  & Observer & Position parse error \\
\texttt{ZState}          & \texttt{object\_states}     & Observer & State decoding error \\
\texttt{ZState}          & \texttt{sprite\_overrides}  & Observer & Per-step appearance error \\
\texttt{ZState}          & \texttt{hud\_values}        & Observer & HUD readout error \\
\texttt{ZState}          & \texttt{event\_objects}     & Observer & Event encoding error \\
\texttt{HState}          & \texttt{object\_velocities} & Simulator & Velocity prediction error \\
\texttt{HState}          & \texttt{event\_triggers}    & Simulator & Trigger statistic error \\
\texttt{HState}          & \texttt{game\_state}        & Simulator & State-machine error \\
\texttt{LevelConstants}  & all fields                  & Observer & Pixel-accuracy regression \\
\texttt{Metadata}        & all fields                  & Harness  & (read-only) \\
\bottomrule
\end{tabular}
\end{table}

\paragraph{Forbidden overlaps.}
A field belongs to exactly one schema. The Observer writes \texttt{ZState} and \texttt{LevelConstants} but reads \texttt{HState}; the Simulator writes \texttt{HState} but reads \texttt{ZState}. A patch that moves an Observer-owned field into \texttt{HState} (for example, copying \texttt{object\_positions} into the dynamics state) is rejected because it duplicates the source of truth and lets the two schemas drift. The same constraint excludes Simulator-only state from \texttt{ZState}: velocities, action effects, and state-machine variables remain in \texttt{HState} or in code-level constants such as \texttt{LEARNED\_EFFECTS}.

\paragraph{Tolerated overlaps.}
Two narrow overlaps are tolerated. First, entity \emph{names} are shared across \texttt{ZState.object\_positions}, \texttt{HState.object\_velocities}, and \texttt{LevelConstants.objects}. The same canonical name keys all three so that a position, a velocity, and a sprite refer to the same entity. Second, \texttt{event\_objects} appears in \texttt{ZState} as a per-step transient and \texttt{event\_triggers} appears in \texttt{HState} as an aggregate statistic over those transients. The Observer encodes the per-step event; the Simulator accumulates the statistic. The two are tied by entity name and action, and either can be edited without touching the other.

\paragraph{Disjointness and localized repair.}
Disjointness makes the diff $e_t = \operatorname{diff}(\hat z_{t+1}, z_{t+1})$ routable. A field-level mismatch on \texttt{ZState} is owned by the Observer when the encoding of $z_{t+1}$ disagrees with the new frame, or by the Simulator when the encoding agrees but \texttt{predict} did not project to it; a sub-goal disagreement under an agreed \texttt{ZState} is owned by the Inductive Explorer. The Critic intervenes only when more than one candidate fits a single failure.

\subsection{Regression checks and retained evidence}
\label{sec:regression-checks}

The regression set \(\mathcal{R}_t\) referenced in Sections~\ref{sec:workspace-optimization}~and~\ref{sec:dreamteam} is not a separate test database; it is a window of past transitions that the harness rebuilds on every step by replaying the current programs against the stored trace. This appendix gives its data shape; Appendix~\ref{app:execution-details} describes how the window is selected and how individual replays are orchestrated.

\paragraph{Per-step retrodiction record.}
Each entry in \(\mathcal{R}_t\) is a structured record produced by replaying the current \texttt{predict}, \texttt{render}, and \texttt{render\_event} on a stored transition $(z_{k-1}, h_k, a_{k-1}, z_k)$ for some $k \le t$. The record is shaped as follows.

\begin{lstlisting}[basicstyle=\ttfamily\small, frame=single, xleftmargin=1em]
@dataclass
class StepRetrodiction:
    step:   int
    level:  int
    action: str
    source: str  # "recent" | "event" | "error_z" | "error_render"

    # predict(z[k-1], h[k], action, ...) vs z[k]
    z_predicted:  dict | None
    z_actual:     dict | None
    z_accuracy:   str | None        # "N/M" matched fields
    z_mismatches: list | None       # per-field counterexamples

    # render(z[k], constants) vs actual_grid[k]
    render_z_accuracy: float | None
    render_z_metrics:  dict | None  # {pixel_accuracy,
                                    #  structural_similarity,
                                    #  object_iou,
                                    #  color_histogram_distance}

    # render(predict(z[k-1], h[k], action, ...), constants) vs actual_grid[k]
    render_pred_accuracy:      float | None
    render_pred_by_hypothesis: dict   # name -> accuracy

    # render_event(z[k], constants) vs actual event frames at step k
    event_accuracy:  float | None
    event_agreement: str | None      # "match" | "missed"
                                      # | "false_positive" | "agree_none"

    # per-hypothesis retrodiction for HYPOTHESES table
    hypothesis_results: dict | None  # name -> {accuracy, mismatches,
                                      #         probability}

    # delta vs previous retrodiction of the same step
    improved:          bool | None
    improvement_detail: str | None
\end{lstlisting}

\paragraph{Window composition.}
The harness composes \(\mathcal{R}_t\) from four selectors with fixed quotas: \emph{recent} (last few transitions), \emph{event} (transitions with events), \emph{error\_z} (worst per-field prediction errors), and \emph{error\_render} (worst pixel-accuracy errors). The pools are unioned and deduplicated, then rebuilt against the patched programs on every step, so a patch that improves a recent entry but breaks an older one surfaces immediately.

\paragraph{Accuracy metrics.}
Three families of metric appear in a record. Field-level accuracy compares \texttt{ZState} dictionaries entry by entry, yielding a matched-fields fraction and named mismatches. Pixel-level accuracy compares two \texttt{64$\times$64} grids on four scores: raw pixel agreement, structural similarity, object-level IoU after instance matching, and color histogram distance. Event-level accuracy yields a categorical agreement label and a per-frame pixel score. The Observer reads pixel and event metrics, the Simulator reads field and per-hypothesis metrics, and the Critic reads the improvement delta.

\paragraph{Replay feedback against \(\mathcal{R}_t\).}
Patches enter the workspace once their source passes a syntactic and exports check (the harness validates that the file loads and that its named exports are present and callable). There is no automatic acceptance gate against \(\mathcal{R}_t\) and no rollback. The replay produces a per-record delta that the harness reports as feedback; records that the patch breaks become new counterexamples for the same owner on the next step. Repeated failures lengthen the entry's footprint on the persistent ledger, which is the signal the Team Leader uses to switch from planning to probing.

\subsection{\textsc{DreamTeam} agentic workspace details}
\label{sec:agentic-workspace-details}

The trainable object in \textsc{DreamTeam} is the workspace, not only the executable world model. The world model is the most visible part of the workspace because it is runnable and regression-tested, but the agent also trains persistent role context around that code: logs, hypotheses, goals, probes, critique notes, routing decisions, action commitments, strategy notes, and retained regression checks. A future model call therefore reads both programs and the structured evidence that explains why those programs have their current form.

We separate the workspace into executable artifacts and role-specific context.

\begin{description}
  \item[Observation/world model.] The observation artifact, implemented as \texttt{observable.py}, defines the structured observation schema and the renderers used to check a parse, not a reusable raw-grid parser. It contains \texttt{ZState}, entity labels, rendering helpers, and optional feature extractors. The Observer reads the latest screen, previous parsed state, parse failures, and reconstruction checks, then emits a per-step \(z_t\) record transductively. It may update entity definitions, schema fields, render helpers, and notes about ambiguous pixels or sprites. For example, after an unexpected reset caused by touching a newly seen object, the Observer may add an \texttt{obstacle} or \texttt{hazard} entity with a shape signature and a reconstruction test over the observed frame. The update enters the workspace once the artifact loads; replay and render checks then report whether it explains the current observation and whether older retained frames now regress. Any regression is surfaced as a counterexample for the next Observer update rather than causing an automatic rollback.

  \item[Dynamics/simulator.] The dynamics artifact, implemented as \texttt{dynamics.py}, defines the predicted effect of actions on structured state. The Simulator reads \texttt{z\_prev}, the committed action, the observed \texttt{z\_next}, stored one-step predictions, and dynamics failure reports. It writes transition rules, constants, preconditions, and rollout checks. For example, a failed prediction that expected ordinary movement but observed death can become a lethal-contact rule guarded by the newly named entity type. Once the patch loads, the harness reports whether it fixes the failed transition and what happens on the retained regression window over previous movement, wall collision, button, reset, and inventory transitions. If the rule is plausible but underdetermined, it remains in the simulator notes as a candidate rule and the Transductive Explorer may be asked for a targeted probe.

  \item[Strategy/inductive explorer.] The strategy artifact, implemented as \texttt{strategy.py}, defines reusable policies, subgoals, and progress checks. The Inductive Explorer reads the current observation state, simulator rollouts, known goals, failed tactics, and the Policies $\times$ Sub-Goals matrix. It writes named policies, goal predicates, rollout summaries, strategy notes, and regression checks for policy progress. For example, once the simulator can predict movement around a lethal object, the Inductive Explorer may add a \texttt{route\_around\_hazard} policy with a progress check requiring reduced distance to a target without entering lethal contact. Simulated rollouts and recent real evidence determine whether the policy is promoted into the reusable strategy library; if it succeeds only in the current trace, it may be retained as a local plan instead.
\end{description}

The non-executable role context is also part of the trainable workspace. It changes what later calls attend to, which actions are considered informative, and which failures are routed to code owners.

\begin{description}
  \item[Transductive Explorer context.] The Transductive Explorer owns local experimentation rather than a durable code artifact. It reads uncertainty tags, failed predictions, unexplained observations, the action budget, and the Team Leader's recent commitments. It writes probe proposals, hypotheses tagged \texttt{[UNTESTED]}, \texttt{[UNCERTAIN]}, or \texttt{[CONFIRMED]}, expected probe outcomes, and stop criteria. A concrete write might be: ``Press the blue tile once; if the door opens, route to dynamics as a toggle rule; if only the sprite changes, route to observation; if neither changes, mark the tile non-interactive.'' Probe outcomes are accepted into the retained workspace when the environment result distinguishes hypotheses or creates a counterexample for an owner. Probes that do not disambiguate are kept as negative evidence so they are not repeated without a new reason.

  \item[Critic context.] The Critic owns failure localization and quality control. It reads role logs, committed predictions, observed outcomes, proposed patches, action plans, and regression deltas. It writes critique notes, failure routing criteria, owner assignments, and blocking objections for the Team Leader and artifact owners. For example, when a patch adds a broad ``all colored blocks are lethal'' rule, the Critic may point to an earlier safe colored block and argue that the next commitment should be delayed until the rule is narrowed. Its updates are retained as routing constraints and regression targets: a note that a failure is parser-owned, simulator-owned, or strategy-owned changes which artifact receives the next patch request. A Critic objection is authoritative only when it is grounded in a logged trace, violated commitment, or failed check; otherwise it remains as a warning rather than a blocker.

  \item[Team Leader context.] The Team Leader owns action commitment and arbitration. It reads the Observer's parse, Simulator predictions, Inductive Explorer rollout results, Transductive Explorer probes, Critic warnings, action budget, and recent success or failure signals. It writes the selected action, the reason for the action, the prediction it is committing the workspace to, and the decision source: probe, rollout-supported plan, reusable policy, repair action, or fallback. For example, it may commit to \texttt{RIGHT} with the stored prediction that the avatar moves one cell right and no reset occurs. After the environment responds, that commitment becomes training evidence. If the prediction holds, the commitment is retained as support for the current parser and simulator; if it fails, the mismatch becomes a routed counterexample rather than an unstructured mistake.

  \item[Inductive Explorer context.] In addition to owning \texttt{strategy.py}, the Inductive Explorer maintains notes that are not yet executable strategy. It reads Team Leader decisions, real progress signals, simulator rollout disagreement, and Critic warnings about degenerate policies. It writes candidate goals, abandoned tactics, rollout caveats, and promotion criteria for turning a local plan into a reusable policy. A concrete retained note might say that a key-collection route worked only after a door-toggle rule was added, so future policy reuse must require the door state to be known. Such notes are accepted when they summarize logged rollouts or real transitions, and they remain non-executable until converted into a policy or regression check.
\end{description}

Durability is deliberately conservative. Executable patches enter \(W_{t+1}\) once the harness confirms they load and that their named exports are present, and the harness then reports their replay fitness against the current counterexample and against \(\mathcal{R}_t\) as per-record feedback. Non-executable context is retained when it is tied to a trace, a prediction, a probe outcome, a routing decision, or an action commitment. Unsupported claims can still remain in the workspace, but only with an uncertainty tag and without the authority to block a patch or justify a plan. In this sense the workspace learns at two levels: programs change the behavior of parsing, simulation, and policy rollout, while role context changes what evidence is collected, how failures are assigned, and which executable edits the role is willing to commit to next.

\section{Execution Details}
\label{app:execution-details}

This appendix contains implementation details deferred from Sections~4 and~5. Section~\ref{sec:multi-action-modes} describes how the Team Leader's per-step commitment is executed when it spans more than one environment action.

\subsection{Multi-Action Execution Modes}
\label{sec:multi-action-modes}

The deliberation step described in Algorithm~\ref{alg:dreamteam-step} commits a single action $a_t$. In practice, the Team Leader can commit a short batch of actions in one response. The harness then executes the batch under a configurable mode that controls how much intermediate computation runs between consecutive actions. The configuration ceiling is \texttt{max\_actions\_per\_step}, with a default of twenty, and the Team Leader's response is parsed by extracting the action sequence from the first non-empty line of its structured output.

Three execution modes are supported. They differ in how the world model and the Team Leader are updated between actions, and in how early-stop triggers interrupt the batch.

\paragraph{Simple mode.}
Every action in the batch is executed against the environment without any intermediate language-model call. The Observer and Simulator do not run between actions; the harness collects observations $o_{t+1}, \dots, o_{t+n}$, appends them to the trajectory, and lets the next deliberation step process the entire batch retroactively. Simple mode is selected when the world model holds high retrodiction accuracy on recent transitions and the action sequence consists of repeated movement in a region the parser already covers. Its purpose is throughput: a confident traversal of a known corridor avoids paying the language-model cost of a per-step deliberation.

\paragraph{Approval-based mode.}
After each action, the Observer and Simulator run lightweight micro-tasks: the Observer encodes the new screen into $z_{t+i+1}$, and the Simulator updates $h_{t+i+1}$ and runs retrodiction on the just-observed transition. The Team Leader then reads the freshly computed state and chooses one of three outcomes for the next action in the batch: \emph{approve} the next action and continue, \emph{reject} the remainder of the batch and return to a full deliberation step, or \emph{request a reset}. Approval-based mode lets the Team Leader stop a planned sequence at the first sign that the world model's prediction has drifted, while still amortizing the cost of the upstream Inductive Explorer rollout that produced the batch.

\paragraph{Policy-driven mode.}
The Team Leader commits to a registered policy and a length, written as \texttt{POLICY:<name>:<n>}. The harness invokes the policy on $(z_{t+i}, h_{t+i}, \dots)$ to choose action $a_{t+i}$ for $i = 0, \dots, n-1$, runs Observer and Simulator micro-tasks after each action so the policy reads up-to-date state, and stops if the policy fails to produce a valid action. The Team Leader is consulted periodically rather than between every action; this lets a state-dependent policy from \texttt{strategy.py} carry the trajectory through a long routine while keeping the world model grounded.

\paragraph{$z_{\text{prev}}$ chaining.}
In every mode that runs micro-tasks between actions, the Observer's encoding from the just-executed action is the $z_{\text{prev}}$ for the next action's prediction and rollout. The chain is propagated through the harness rather than reconstructed from scratch, so a stale $z_{\text{prev}}$ from before the batch started cannot leak into a later prediction. This makes multi-action retrodiction equivalent, action by action, to the single-action retrodiction described in Section~\ref{sec:step-loop}: each action in the batch produces its own diff $e_{t+i}$ that is routed to the same owners by the same disjointness rule.

\paragraph{Early-stop triggers.}
A batch in approval-based or policy-driven mode is interrupted by any of the following conditions, evaluated after each action: a level-completion increment in the harness's \texttt{levels\_completed} counter; a non-empty transient event captured by the Observer's \texttt{event\_objects} encoding; a \texttt{GAME\_OVER} return from the environment; or a reset request raised by the Team Leader's approval gate. Each of these conditions either changes the workspace in a way that invalidates the precomputed batch (a level transition introduces new \texttt{LevelConstants}; an unexpected event surfaces a counterexample to the dynamics model) or signals that the agent should re-deliberate.

\paragraph{Crash recovery and the preferred action.}
The harness records the proposed batch at the start of execution. When an action mid-way through the batch raises an exception or times out, the next deliberation step starts with a \emph{preferred action} hint set to the first unplayed action in the batch. The next Team Leader decision is biased toward continuing the planned sequence rather than restarting from scratch, so a transient failure in the middle of an approved trajectory does not cost the rest of the trajectory.

\paragraph{Action accounting.}
Multi-action is the only mechanism that produces more than one environment transition per deliberation step, and is therefore the dominant lever on the per-action cost reported in Section~\ref{sec:experiments}. The action budget against the human baseline counts environment transitions, so multi-action changes the language-model cost charged to each transition without changing the denominator.

\section{Additional Experiments}
\label{app:additional-experiments}

This appendix records the experimental setup for the runs analysed in
Section~\ref{sec:experiments}, the per-run reproducibility breakdown,
and the benchmark-version caveat needed to interpret comparisons
against published ARC-AGI-3 numbers. Ablations, synthetic-game
results, and game-family stratification are not claimed here.

\subsection{Experimental setup}
\label{app:experimental-setup}

Table~\ref{tab:app-hparams} records the per-role model
assignment and the team-wide hyperparameters of the run
analysed in Section~\ref{sec:experiments}.

\begin{table}[htbp]
\centering
\small
\caption{Experimental setup. \textsc{DreamTeam} on ARC-AGI-3
public set, competition mode, single seed.}
\label{tab:app-hparams}
\begin{tabular}{ll}
\toprule
\textbf{Parameter} & \textbf{Value} \\
\midrule
Operation mode               & competition \\
Games                        & 25 (public set) \\
Seed                         & 0 \\
Dedicated step-wall cap      & none in the analysed run \\
Per-agent deliberation cap   & $300$~s \\
LLM-call timeout             & $180$~s \\
Team-leader timeout          & $300$~s \\
Max actions per step (MA)    & $20$ \\
Reset cooldown               & $5$ steps \\
Safety-round depth           & $4$ \\
Safety-round breadth         & $6$ \\
Per-step token budget        & $250{,}000$ tokens \\
Context window (last-$k$)    & $6$ rounds, $\le 500$K tokens \\
Reasoning compaction         & strip after $1$ turn \\
Summarization                & on (\textsc{flash}, $\le 50$K tokens) \\
Policy-action execution      & enabled \\
Reset reflection             & enabled \\
DSL injection in prompts     & off \\
Inter-role chat channel      & off (file-based comms only) \\
\midrule
\multicolumn{2}{l}{\textit{Per-role models}} \\
Observer / Simulator         & Claude Opus 4.6 (no reasoning) \\
Team Leader / TE             & Claude Opus 4.6 (low reasoning) \\
Inductive Explorer / Critic  & GPT-5.5 (high reasoning) \\
\bottomrule
\end{tabular}
\end{table}

\subsection{Run-to-run reproducibility}
\label{app:second-run}

The headline numbers in Section~\ref{sec:experiments} average over
two independent runs of \textsc{DreamTeam} on the same 25 games
with identical hyperparameters (Table~\ref{tab:app-hparams}). Run~1
completed all 25 games within the protocol budget; run~2 was
operator-stopped while 18 games were still in progress. For those
18 games we read the \emph{partial-run RHAE} from per-step
\texttt{level\_score} events through the run-2 stop point; the
formula is the same as the official scorer
(Eq.~\ref{eq:rhae-level}--\ref{eq:rhae-game}). Table~\ref{tab:app-pairwise}
shows the pairwise per-game RHAE on the seven games that completed
in both runs.

\begin{table}[htbp]
  \caption{Per-game RHAE (\%) on the seven games that completed in
  both runs. Mean delta across the seven (run~2 $-$ run~1):
  $-0.20$~pp, median $0$. Notable per-game movements: R11L (run~2
  cleared L4--L5 that run~1 missed) and SB26 (run~2 regressed on
  later levels) cancel each other out.}
  \label{tab:app-pairwise}
  \centering
  \small
  \begin{tabular}{lrrr}
    \toprule
    \textbf{Game} & \textbf{Run 1} & \textbf{Run 2} & \textbf{Per-game mean} \\
    \midrule
    AR25 & 100.00 & 100.00 & 100.00 \\
    CD82 &  79.05 &  90.41 &  84.73 \\
    FT09 & 100.00 &  92.24 &  96.12 \\
    LP85 & 100.00 & 100.00 & 100.00 \\
    R11L &  47.62 &  88.62 &  68.12 \\
    SB26 &  93.96 &  47.97 &  70.97 \\
    TU93 & 100.00 & 100.00 & 100.00 \\
    \bottomrule
  \end{tabular}
\end{table}

Aggregated across all 25 games, run~1 alone reaches mean RHAE
$38.06\%$ on $9{,}789$ env actions and run~2 reaches $38.67\%$ on
$12{,}391$ env actions (the latter using partial-run RHAE for the
18 games not yet finalised when run~2 was stopped). The per-game
mean of the two runs is $\mathbf{38.36\%}$ at a mean of $\mathbf{444}$
env actions per game; this is the number reported in
Section~\ref{sec:experiments}. The per-step pooled denominator
($22{,}180$ env actions across the two runs) is what
Table~\ref{tab:exp-per-role} uses for per-role averages.

\subsection{RHAE scoring formula}
\label{app:rhae-formula}

The official ARC-AGI-3 scorer\footnote{\url{https://docs.arcprize.org/methodology}, accessed May~7, 2026.} computes RHAE in three stages. Let \(B_{g,\ell}\) be the human-baseline action count for game~\(g\), level~\(\ell\), and \(a_{g,\ell}\) the agent's action count on that level. The per-level score is
\begin{equation}
\mathrm{RHAE}_\ell(g) \;=\; \min\!\left(1.15,\; \left(\frac{B_{g,\ell}}{a_{g,\ell}}\right)^{2}\right),
\label{eq:rhae-level}
\end{equation}
i.e., the squared action ratio capped at \(1.15\). The per-game score is the level-weighted mean using level-index weights \(w_\ell = \ell\) (1-indexed), bounded by the fraction of weights for levels actually solved:
\begin{equation}
\mathrm{RHAE}(g) \;=\; \frac{\sum_{\ell \in \mathrm{solved}(g)} \ell \cdot \mathrm{RHAE}_\ell(g)}{\sum_{\ell=1}^{L_g} \ell},
\label{eq:rhae-game}
\end{equation}
where \(L_g\) is the total number of levels in game \(g\). Unsolved levels contribute zero to the numerator while still counting in the denominator, so completion is a hard ceiling: a game cannot reach 100\% without all levels cleared. The headline number is the unweighted mean of \(\mathrm{RHAE}(g)\) over the 25 public games.

\subsection{Benchmark version shift and human-baseline drift}
\label{app:benchmark-version-shift}

The ARC-AGI-3 changelog for April~14, 2026 changed the public scoring methodology and republished fifteen game versions.\footnote{\url{https://docs.arcprize.org/changelog}, accessed May~5, 2026.} The scoring change is itself important: the human baseline now uses the median human per level rather than the second-best human, and the per-level score cap increased from \(1.0\times\) to \(1.15\times\) the human baseline. Comparisons across this boundary therefore conflate agent quality, benchmark content, and scoring protocol. We consequently pin every experiment to exact game hashes and treat leaderboard numbers from the older versions as contextual rather than protocol-matched evidence.

To quantify the direction of the content shift, let \(B_{g,\ell}\) denote the published human-baseline action count for game \(g\) and level \(\ell\), and let \(H_g=\sum_\ell B_{g,\ell}\). We use \(H_g\) as a benchmark-native proxy for action-efficiency pressure: larger values indicate a longer median-human solution and therefore a larger action budget under normalized scoring. This is not a semantic difficulty metric. A game can become easier for humans but harder for an agent if the changed mechanic stresses perception, memory, or exploration. It is nevertheless the budget signal used by the benchmark and by our reset thresholds.

Across the fifteen republished games, the aggregate baseline increased from \(9{,}215\) to \(10{,}253\) actions, a net increase of \(1{,}038\) actions or \(11.3\%\) when weighted by old baseline size. Averaging the per-game relative changes gives a larger \(15.8\%\) increase, because several smaller games were revised upward substantially. Eleven of the fifteen games became harder by more than two percent; restricted to those games, the old and new totals are \(6{,}597\) and \(7{,}926\), a \(20.1\%\) weighted increase and a \(25.7\%\) mean per-game increase. Two games became easier, and two are effectively unchanged by total baseline actions.

\begin{table}[htbp]
  \caption{Aggregate shift in human-baseline action budgets for the fifteen ARC-AGI-3 games republished on April~14, 2026. Weighted relative change is computed from total old and new baseline actions; mean per-game change averages game-level percentage changes.}
  \label{tab:benchmark-version-shift-aggregate}
  \centering
  \small
  \begin{tabular}{lrrrrr}
    \toprule
    \textbf{Subset} & \textbf{Games} & \textbf{Old total} & \textbf{New total} & \textbf{Weighted rel.} & \textbf{Mean rel.} \\
    \midrule
    All republished games & 15 & 9{,}215 & 10{,}253 & +11.3\% & +15.8\% \\
    Harder games only & 11 & 6{,}597 & 7{,}926 & +20.1\% & +25.7\% \\
    Easier games only & 2 & 1{,}392 & 1{,}091 & -21.6\% & -24.0\% \\
    Near-ties & 2 & 1{,}226 & 1{,}236 & +0.8\% & +0.6\% \\
    \bottomrule
  \end{tabular}
\end{table}

\begin{table}[htbp]
  \caption{ARC-AGI-3 version shift on April~14, 2026, measured by total human baseline actions across levels. Larger totals indicate a longer median-human solution and are treated as harder for budget-normalized evaluation. A two-percent deadband is labeled near-tie.}
  \label{tab:benchmark-version-shift}
  \centering
  \small
  \begin{tabular}{lrrrrl}
    \toprule
    \textbf{Game} & \textbf{Old total} & \textbf{New total} & \textbf{Delta} & \textbf{Rel.} & \textbf{Direction} \\
    \midrule
    SC25 & 216 & 350 & +134 & +62\% & Harder \\
    SK48 & 696 & 1070 & +374 & +54\% & Harder \\
    R11L & 167 & 233 & +66 & +40\% & Harder \\
    AR25 & 577 & 748 & +171 & +30\% & Harder \\
    TN36 & 250 & 317 & +67 & +27\% & Harder \\
    TU93 & 378 & 462 & +84 & +22\% & Harder \\
    RE86 & 1071 & 1255 & +184 & +17\% & Harder \\
    M0R0 & 970 & 1107 & +137 & +14\% & Harder \\
    SP80 & 472 & 518 & +46 & +10\% & Harder \\
    S5I5 & 608 & 638 & +30 & +5\% & Harder \\
    DC22 & 1192 & 1228 & +36 & +3\% & Harder \\
    CN04 & 779 & 789 & +10 & +1\% & Near-tie \\
    VC33 & 447 & 447 & 0 & 0\% & Near-tie \\
    KA59 & 826 & 730 & -96 & -12\% & Easier \\
    SU15 & 566 & 361 & -205 & -36\% & Easier \\
    \bottomrule
  \end{tabular}
\end{table}

Table~\ref{tab:benchmark-version-shift} shows that the largest upward revisions are SC25, SK48, R11L, AR25, and TN36. KA59 and SU15 became easier by this proxy. VC33 is unchanged, and CN04 is a near-tie by total baseline actions despite a substantial content change: it moves from five to six levels, adds per-level step limits, and changes later-level masking/background structure. These mixed cases are the main reason we report exact game hashes rather than only game names.

% ── NeurIPS Checklist ────────────────────────────────────────────────
% \newpage
% \input{checklist.tex}

\end{document}